\documentclass{article}

\PassOptionsToPackage{numbers, compress}{natbib}

\usepackage{bm}

\usepackage[preprint]{neurips_2022}

\usepackage[utf8]{inputenc} 
\usepackage[T1]{fontenc}    
\usepackage{hyperref}       
\usepackage{url}            
\usepackage{booktabs}       
\usepackage{amsfonts}       
\usepackage{nicefrac}       
\usepackage{microtype}      
\usepackage{xcolor}         
\usepackage{amsmath,amssymb}
\usepackage{tikz}
\usepackage{comment}
\usepackage{todonotes}
\usepackage{multirow}
\usepackage{color}

\title{Coupling Visual Semantics of Artificial Neural Networks and Human Brain Function via Synchronized Activations}

\author{%
Lin Zhao \\
University of Georgia\\
\texttt{lin.zhao@uga.edu}
\And
Haixing Dai \\
University of Georgia\\
\texttt{haixing.dai@uga.edu}
\And
Zihao Wu \\
University of Georgia\\
\texttt{zihao.wu1@uga.edu}
\And
Zhenxiang Xiao \\
University of Electronic Science and Technology of China \\
\texttt{zhenxiang.up@gmail.com}
\And
Lu Zhang \\
University of Texas at Arlington \\
\texttt{lu.zhang2@mavs.uta.edu}
\And
David Weizhong Liu \\
Athens Academy \\
\texttt{david.weizhong.liu@gmail.com}
\And
Xintao Hu \\
Northwestern Polytechnical University \\
\texttt{xhu@nwpu.edu.cn} 
\And
Xi Jiang \\
University of Electronic Science and Technology of China \\
\texttt{xijiang@uestc.edu.cn}
\And
Sheng Li \\
University of Georgia\\
\texttt{sheng.li@uga.edu} \\
\And
Dajiang Zhu \\
University of Texas at Arlington \\
\texttt{dajiang.zhu@uta.edu}
\And
Tianming Liu \\
University of Georgia\\
\texttt{tliu@uga.edu} \\
}

\begin{document}

\maketitle

\begin{abstract}
Artificial neural networks (ANNs), originally inspired by biological neural networks (BNNs), have achieved remarkable successes in many tasks such as visual representation learning. However, whether there exists semantic correlations/connections between the visual representations in ANNs and those in BNNs remains largely unexplored due to both the lack of an effective tool to link and couple two different domains, and the lack of a general and effective framework of representing the visual semantics in BNNs such as human functional brain networks (FBNs). To answer this question, we propose a novel computational framework, Synchronized Activations (Sync-ACT), to couple the visual representation spaces and semantics between ANNs and BNNs in human brain based on naturalistic functional magnetic resonance imaging (nfMRI) data. With this approach, we are able to semantically annotate the neurons in ANNs with biologically meaningful description derived from human brain imaging for the first time. We evaluated the Sync-ACT framework on two publicly available movie-watching nfMRI datasets. The experiments demonstrate a) the significant correlation and similarity of the semantics between the visual representations in FBNs and those in a variety of convolutional neural networks (CNNs) models; b) the close relationship between CNN's visual representation similarity to BNNs and its performance in image classification tasks. Overall, our study introduces a general and effective paradigm to couple the ANNs and BNNs and provides novel insights for future studies such as brain-inspired artificial intelligence.
\end{abstract}

\section{Introduction}

Inspired by the biological neural networks (BNNs), artificial neural networks (ANNs) have achieved great success in a variety of tasks and scenarios due to their powerful representation ability~\cite{lecun2015deep}. In computer vision (CV) field, convolutional neural networks (CNNs)~\cite{lecun1995convolutional} hierarchically learn the visual representations of images/videos as low-level to high-level features in embedding space and have been widely used in many real-word applications~\cite{khan2020survey,lecun2015deep}. Recent Vision Transformer (ViT) \cite{dosovitskiy2020image} demonstrates promising performance by representing the image as a sequence of patches and embedding the image patches as latent vectors, based on which the dependencies/correlations among those vectors are modeled. However, the semantics of those embedding spaces for visual representation are not manifest for human perception and challenge us for a comprehensive understanding of representation learning of ANNs. To unveil and describe the semantics of latent space of ANNs for visual representation, increasing efforts have been devoted to interpret the ANNs' behaviors and annotating their neurons with semantic concepts~\cite{zhou2016learning,bau2017network}. For example, Bau et al.~\cite{bau2017network} proposed to label the hidden units of convolutional layer with visual concepts from a broad dataset. A recent study employed fine-grained natural language description to annotate the semantics of neurons in various ANNs~\cite{hernandez2021natural}. Despite the remarkable progresses achieved by these methods, whether the visual representation space of ANNs retains biologically meaningful semantics as in the initial inspiration, BNNs, is still an open question.

In the field, researchers now have employed naturalistic functional magnetic resonance imaging (nfMRI) to assess the activity and functional mechanism of BNNs~\cite{hu2010bridging,liu2014merging,wang2017test,ren2021hierarchical}, e.g., functional brain networks (FBNs), under the naturalistic stimuli such as real-life images and video streams. This natural stimulus fMRI paradigm provides a powerful tool for investigating the visual perception of human brain and representing the corresponding semantics~\cite{liu2014merging}, allowing us to answer the aforementioned question and annotate the neurons in ANNs with biological description even further. However, the current approaches for representing high-dimensional fMRI data, e.g., matrix decomposition based on independent component analysis (ICA)~\cite{calhoun2006unmixing} and sparse dictionary learning (SDL) \cite{lv2014holistic}, are commonly used for task and resting state fMRI. Considering the brain activities encoded by nfMRI are dynamic and complex, it is quite challenging to interpret and describe the semantics perceived by the human brain. In addition, the brain responses evoked by naturalistic stimuli exhibit great inter-subjects variability ~\cite{golland2007extrinsic, ren2017inter}, while those existing methods do not encode the regularity and variability of different brains, and thus do not offer a general, comparable, and stereotyped embedding space for representing the brain activity and functional semantics. Recently, deep learning approaches demonstrated superior performance in modeling fMRI data \cite{wang2018recognizing,zhang2019identify,liu2019cerebral,dong2019modeling,zhao2021exploring,li2021evolutional}. However, as far as we know, these deep learning methods were designed for specific tasks. A more general and effective framework of embedding brain function and representing semantics under naturalistic stimuli is still much needed. In parallel, linking such human brain's functional embedding and semantics representation with external natural stimulus is very desirable and significant.  

\begin{figure}[htb]
  \centering
  \includegraphics[width=1.0\linewidth]{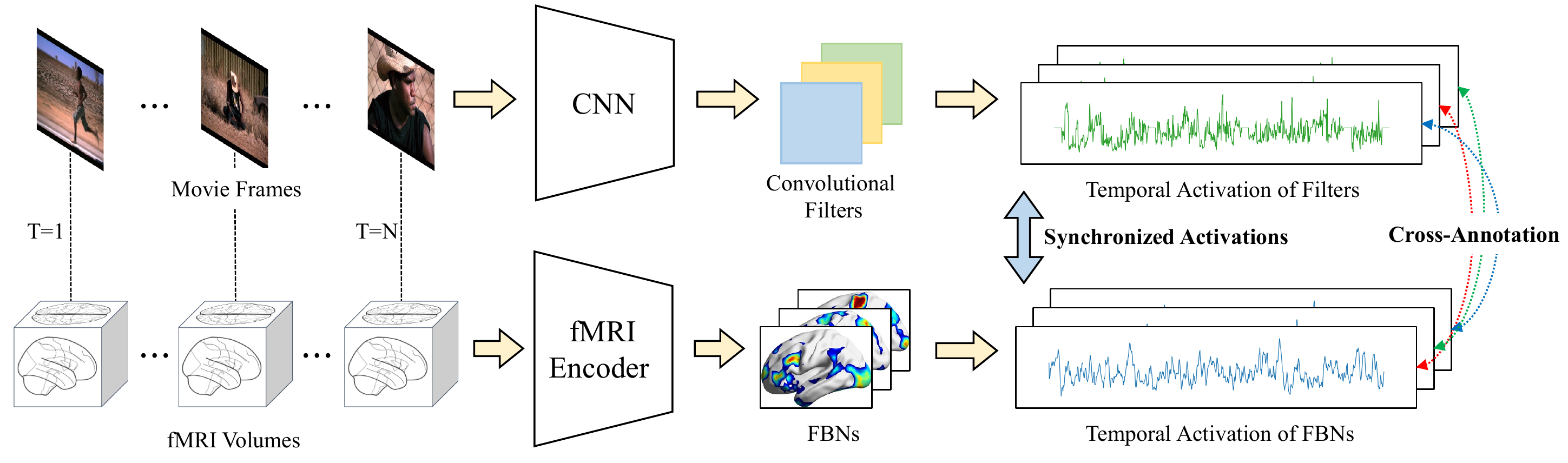}
  \caption{The proposed Sync-ACT framework. The temporal activation of FBNs and convolutional filters are synchronized for matching the embedding space and cross-annotation.}
  \label{figure1}
\end{figure}

In this work, we propose a novel computational framework, Synchronized Activations (Sync-ACT), to explore the connections of the visual representation space and semantics between ANNs and BNNs in human brain. Based on Sync-ACT, we describe and annotate the neurons in ANNs with biologically meaningful descriptions for the first time, bridging the gaps between these two drastically different domains. Specifically, as illustrated in Fig.~\ref{figure1}, we propose an fMRI embedding framework which represents the brain activity during movie watching as FBNs with temporal activations. The frames in each time point of the corresponding movie are input into CNN models and the maximum activation values in features map of each convolutional filter are recorded to form its temporal activation over time. In this way, the activation of convolutional filters and FBNs in human brain are synchronized, which is supported by the fact that video frames and functional brain responses are intrinsically aligned along the temporal axis during nfMRI scan. By correlating the synchronized temporal activation of FBNs and convolutional filters, the embedding spaces of BNNs in the human brain and in CNNs are matched and connected. In this way, the semantics in spaces of two different domains can be used to cross-annotate each other. We evaluate the proposed framework on the publicly available Human Connectome Project (HCP) 7T movie-watching task fMRI dataset~\cite{van2013wu} and StudyForrest movie-watching fMRI dataset \cite{hanke2016studyforrest}. The experiments demonstrate significant correlations and the similarity of semantics between the two embedding spaces in a variety of CNN models, suggesting a new paradigm for the interpretability studies of both ANNs and BNNs. It is also found that the similarity of visual representations in CNNs to those in BNNs is closely related to its performances in image classification tasks, which could be potentially utilized for guiding the design of ANNs, i.e., for neural architecture search and brain-inspired artificial intelligence. 

\section{Related Works}

\subsection{Visual Representation Interpretation}
Interpreting the behavior of deep neural networks and the learned visual representation has attracted growing interests in the CV field. As the semantics of visual representation in deep networks are not manifest for human perception, a possible approach is to visualize the activation of neurons and characterize the visual concept they recognize~\cite{bau2017network,dalvi2019one,morcos2018importance,mu2020compositional}. For example, Bau et al.~\cite{bau2017network} labels the neurons (i.e., convolutional filters) of CNNs with visual concepts by aligning the activation of neurons with a set of images with semantic concepts. Mu \& Andreas~\cite{mu2020compositional} searches the compositional logical concepts defined on primitive visual concepts that closely approximate neuron behavior. A recent study employed fine-grained natural language description to annotate the semantics of neurons in various ANNs~\cite{hernandez2021natural} by maximizing the mutual information between the language description and imaging regions in which the neuron is activated. Our work is inspired by and in line with aforementioned interpretation studies. The neuron's behaviors are measured by the maximum activation over the time, forming a temporal activation series with which the temporal activation of FBNs is compared and correlated for alignment. In this way, we contribute a biologically meaningful description for the neurons in ANNs.

\subsection{fMRI Data Representation}

A major challenge for fMRI data representation learning is that the number of voxels in 4D spatiotemporal fMRI data is greatly larger than the number of subject brains \cite{mwangi2014review}. To deal with this imbalance, a variety computational tools have been proposed to select the task-related features and discard the redundant ones as well as the noises \cite{calhoun2006unmixing,lv2014holistic}. For example, independent component analysis (ICA) \cite{calhoun2006unmixing} and sparse dictionary learning (SDL) were employed to decompose the fMRI as two compact matrices (temporal and spatial patterns). However, the temporal and/or spatial patterns obtained from ICA or SDL based methods are not intrinsically comparable across different individual brains. Recently, deep learning has been widely employed in fMRI data modeling and achieved superior results over the traditional matrix decomposition methods ~\cite{wang2018recognizing,zhang2019identify,liu2019cerebral,dong2019modeling,zhao2021exploring,li2021evolutional}. However, as far as we know, prior deep learning models of fMRI data were not specifically designed towards a general, comparable and compact representation of brain function. Instead, prior methods were designed for some specific tasks, such as fMRI time series classification \cite{liu2019cerebral}, brain network decomposition \cite{dong2019modeling, li2021simultaneous}, brain state differentiation\cite{wang2018recognizing}, among others.
Even though some methods derive comparable temporal patterns~\cite{,li2021simultaneous,li2021evolutional}, which might be suitable to our objective, they still rely on matrix composition to obtain the spatial patterns that are not comparable. In this work, we proposed a more general and unified framework to represent the fMRI data from different subjects as functional brain networks and their temporal activations in a general, comparable, and stereotyped latent space. This design enables us to explore the correlation between the semantics of this latent space and those in CNNs.

\subsection{Connection of ANNs and BNNs}

Current ANNs are inspired by the BNNs at the beginning. For examples, CNNs are inspired by the hierarchical organization of vision system in human's primary vision cortex~\cite{kim2016convolutional}. Recently, there is a growing interest on exploring the potential connections between ANNs and BNNs. For instance, the receptive field analysis reveals that the receptive fields of filters in CNNs become progressively larger~\cite{luo2016understanding} and more complex, which is similar to the ventral pathway in cerebral cortex~\cite{barrett2019analyzing}. The filters in the last convolutional layer have class-specific receptive fields akin to concept-cells in visual cortex~\cite{mahendran2016visualizing}. Yamins et al.~\cite{yamins2016using,yamins2014performance} synthesized CNNs outputs by linear regression to predict the neural responses in both the V4 and inferior temporal (IT) cortex. This shows a strong correlation between a CNN’s categorization performance and its ability to predict individual IT neural responses, implicitly indicating the potential representation similarity between ANNs and BNNs. You et al.~\cite{you2020graph} proposed a graph-based representation of ANNs called relational graph, and found that top-performing ANNs have graph structure similar to those of BNNs. Inspired by these studies, we explore the semantic similarity of visual representations in CNNs and the functional representation of human brain, providing a novel insight on the connection of ANNs and BNNs.

\section{Methods}
\subsection{Formulation of Sync-ACT Framework}
Even though the ANNs are originally inspired by the BNNs, the input/output, operating and reasoning processes of neural networks in two domains are quite different and not comparable. Our intuition is to avoid being trapped by the remarkable differences but focus on their responses, such as the activation of neurons, to the external stimuli. In this way, the behavior of the neural networks measured by the responses (i.e., the temporal activation of neurons) in two domains can be directly compared if the stimuli are synchronized, and thus the most similar neurons in two domains can be easily identified and paired. Let $\mathcal{F}:X_a\rightarrow Y_a$ represents an artificial neural network, and $f_i(\mathbf{x_a})$ represents the temporal activation of neuron $f_i$ with respect to stimulus sequence $\mathbf{x_a}$. Similarly, let $\mathcal{G}:X_b\rightarrow Y_b$ represent a biological neural network, and $g_j(\mathbf{x_b})$ denotes the temporal activation of neuron $g_j$ with the stimulus sequence $\mathbf{x_b}$. If the stimuli $\mathbf{x_a}$ and $\mathbf{x_b}$ are synchronized, the paired neuron of $f_i$ in the biological neural network $\mathcal{G}$ can be defined by: 
\begin{equation}
    \label{eq1}
    {\rm Sync\textit{-}ACT}(f_i,\mathcal{G}) = \mathop{\arg\max}\limits_{g_j\in \mathcal{G}}\:\delta(f_i(\mathbf{x_a}),g_j(\mathbf{x_b})),
\end{equation}
where $\delta(\cdot)$ is the measurement of similarity between two temporal activations. Similarly, we could define the paired neuron of $g_i$ in the artificial neural network $\mathcal{F}$ as:
\begin{equation}
    \label{eq2}
    {\rm Sync\textit{-}ACT}(g_i,\mathcal{F}) = \mathop{\arg\max}\limits_{f_j\in \mathcal{F}}\:\delta(g_i(\mathbf{x_b}),f_j(\mathbf{x_a})).
\end{equation}
In this work, we adopt the Pearson correlation coefficient (PCC) for similarity measurement $\delta(\cdot)$. Based on Eq.~\eqref{eq1} and Eq.~\eqref{eq2}, we can obtain the paired neuron $g_j$/$f_j$ for any $f_i$/$g_i$ by choosing the one with the most significant similarity value. With the neuron pairs, the semantics of one neural network can be used to annotate the other, i.e., the cross-annotation. We define the semantic description of neurons $f_i$/ $g_i$ as $d_{f_i}$/$d_{g_i}$. The cross-annotation of paired neurons is then denoted as $d_{f_i}\rightarrow d_{g_j}$ and $d_{g_i}\rightarrow d_{f_j}$. 

We adopted nfMRI data to evaluate the Sync-ACT framework by leveraging the fact that, during nfMRI scan, video frames (stimuli for both ANNs and BNNs) and functional brain responses measured by fMRI are temporally aligned in an intrinsic fashion. In Section~\ref{fmri_sync}, we detail a feasible approach about how to derive the BNNs from nfMRI, define neurons in BNNs, and introduce how to obtain the corresponding temporal activations. In Section~\ref{cnn_sync}, we use CNNs as the representative ANNs and convolutional filters as neurons, and introduce the way to measure the temporal activations of each filter.

\subsection{fMRI Embedding Framework}
\label{fmri_sync}
In the brain imaging field, it is common to represent the brain function as interactions of FBNs and the corresponding temporal patterns. In this way, the FBNs can be viewed as the neurons of BNN in human brain, and temporal patterns represent the activations of those FBNs, which seems to satisfy our desire. However, the previous methods including the deep learning ones do not offer a general and stereotyped space in modeling FBNs. The FBNs and corresponding temporal activations are not intrinsically comparable across different individual brains. 

So, in this section, we propose a general fMRI embedding framework to represent brain function as FBNs and derive the temporal activations in a unified and comparable embedding space. Specifically, the fMRI embedding framework has an encoder-decoder architecture. Fig.~\ref{figure2} illustrates the major components in the encoder. The rearranged 2D fMRI signal matrix $\mathbf{S}\in \mathbb{R}^{t\times n}$, where $t$ is the number of time points and $n$ is the number of voxels, is firstly embedded as a new feature matrix $\mathbf{S}_f\in \mathbb{R}^{t\times m}$ through a learnable transformation matrix $\mathbf{W}\in \mathbb{R}^{n\times m}$, where $m$ is the reduced feature dimension $(m \ll n)$. This transformation can be viewed as compressing the voxels in 3D volume space into $m$ components, i.e., $m$ functional brain networks, by linear combination. The columns in transformation matrix $\mathbf{W}$ recorded the contributions of the voxels to each FBN, i.e., the composition of each FBN, which can be mapped back to 3D volume space for visualizing the spatial pattern of FBN. It is noted that the linear transformation in the encoder parameterized by $\mathbf{W}$ is optimized in a data-driven manner and consistent for all subjects, which guarantees the comparability of $\mathbf{S}_f$ for all subjects. 

\begin{figure}[ht]
  \centering
  \includegraphics[width=0.9\linewidth]{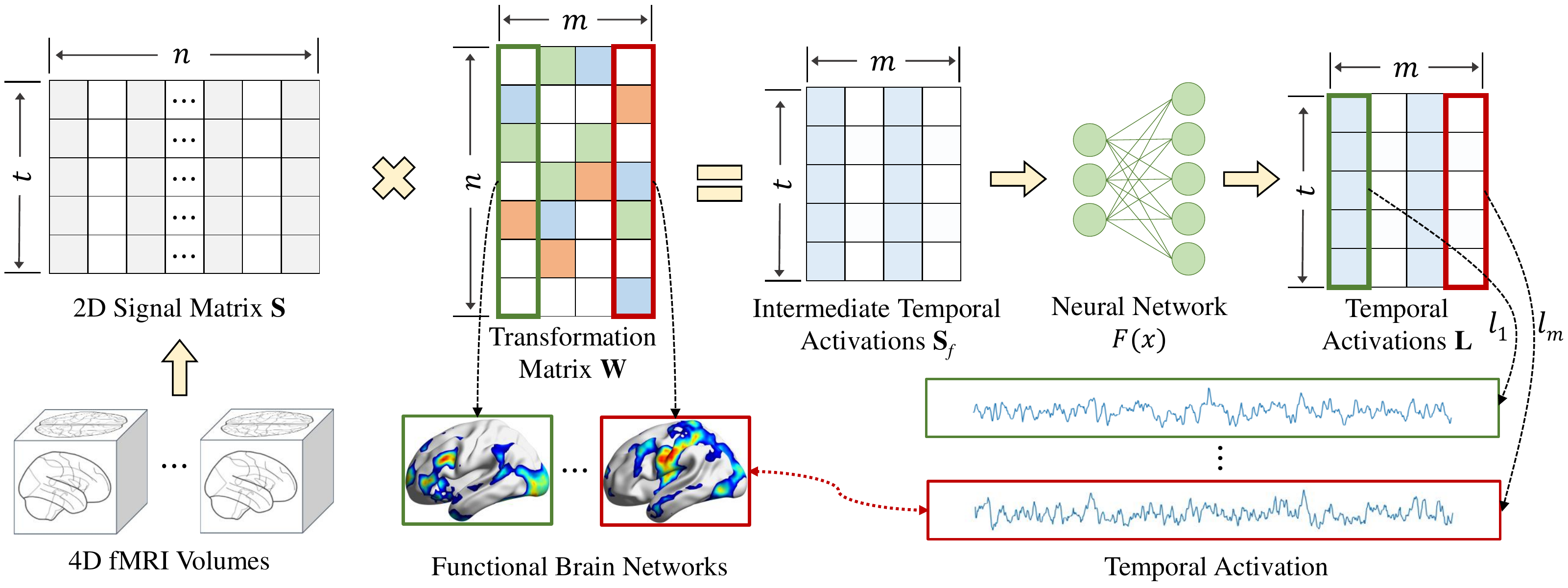}
  \caption{The illustration of the encoder in fMRI embedding framework. The \textcolor{green}{green} and \textcolor{red}{red} boxes correspond to the first and last FBNs and their temporal activations.}
  \label{figure2}
\end{figure}

The row vectors in matrix $\mathbf{S}_f$ recorded the activation of all resulted FBNs at different time points, and the column vector represents the temporal activation of a specific FBNs. We further model the temporal correlations of the column vectors with a neural network $F(x)$. Here, we explore two popular neural networks for modeling temporal data, long short-term memory (LSTM)~\cite{hochreiter1997long} and multi-head self-attention (MSA) module in the Transformer model~\cite{vaswani2017attention}. The column vector $l_i$ in the resulted matrix $\mathbf{L}=F(\mathbf{S}_f), \mathbf{L}\in\mathbb{R}^{t\times m}$ is the temporal activation of the $i^{th}$ FBNs, which encodes the regularity and variability of different brains in the same latent space. We average the vector $l_i$ from all subjects in testing dataset as the temporal activation for neuron $g_i$

The decoder has a symmetrical architecture as the encoder. The whole framework is optimized in an unsupervised manner by minimizing the Mean Square Error (MSE) between the original fMRI signals matrix $\mathbf{S}\in \mathbb{R}^{t\times n}$ and their corresponding reconstruction $\mathbf{S}'\in \mathbb{R}^{t\times n}$.

\subsection{Neurons and Activations in CNNs}
\label{cnn_sync}
We adopt CNNs as the representative ANNs in this work because of CNNs' powerful visual representation ability and wide application in many computer vision tasks. We recognize the convolutional filters as the neurons in ANN. To derive the temporal activation of CNN's filters, we adopt a simple but effective strategy by collecting the feature map $A_{f_i}(x_t)$ of each CNN filter $f_i$ with image $x_t$ from the image sequence $\mathbf{x}$ at time point $t$. Then the maximum value in feature map $\max(A_{f_i}(x_t))$ is extracted to represent the activation degree of filter $f_i$ at time point $t$, resulting in the temporal activation $f_i(\mathbf{x})$. When the image sequence ${\bm{x}}$ is the corresponding movie frames of nfMRI, the derived temporal activations are automatically synchronized with the ones in FBNs. This strategy can be easily applied to any pre-trained CNN models.

With the obtained temporal activations $f_i(\mathbf{x_a})$ and $g_j(\mathbf{x_b})$, given synchronized stimuli $\mathbf{x_a}$ and $\mathbf{x_b}$, we will be able to pair the neurons between ANNs and BNNs and perform cross-annotation following Eq.~\eqref{eq1} and Eq.~\eqref{eq2}. It is noted that the Sync-ACT is a general framework which is also compatible with temporal activations derived from other representation methods, such as those potentially from other fMRI embedding methods or ViT.

\section{Experiments}
\label{experiments}
\textbf{Datasets.} In this study, we adopt the publicly available HCP 7T movie-watching fMRI dataset of S1200 release \cite{barch2013function}. The dataset contains 184 subjects who were scanned in 4 runs while watching short independent film and Hollywood movie excerpts concatenated into .mp4 files. The fMRI data are preprocessed by HCP minimal preprocessing pipeline~\cite{glasser2013minimal}. Then, we downsample and register the preprocessed fMRI data into the standard MNI 152 4 $mm$ space for reducing the computational overhead. In addition, we adopt the StudyForrest movie-watching fMRI dataset \cite{hanke2016studyforrest} with 15 subjects watching 2 hours of Forrest Gump movie. The fMRI data in StudyForrest dataset are preprocessed using fMRIPrep~\cite{esteban2019fmriprep}. It is noted that the data quality, spatial/temporal resolution of the StudyForrest dataset is relatively low and only 15 subjects are available, so we just use it for validation. For both dataset, the time series from the voxels of preprocessed fMRI data are rearranged into a 2D array with zero mean and standard deviation one. More details of data acquisition, fMRI and movie frame preprocessing refer to the supplementary materials.

For both datasets, we used 60\%/10\% of the subjects for model training/validation and the rest 30\% for testing. Unless we specifically mentioned, all the experimental results are based on testing data of HCP 7T movie-watching dataset.

\textbf{Implementation Details.}
In our experiments, we uniformly set the number of derived FBNs from our fMRI embedding framework as 64. The investigation of its influences can be found in supplementary materials. For StudyForrest fMRI dataset, we cut the fMRI data from run \#1 to run \#7 with 430 time points and consider them as the same samples for training due to the lack of subjects. The inference is conducted for each run with the uncut data. The framework is implemented with PyTorch (\url{https://pytorch.org/}) deep learning library. We use the Adam optimizer \cite{kingma2014adam} with $\beta_1=0.9$ and $\beta_2=0.999$. The batch size is 16 and the model is trained for 100 epochs with an initial learning rate 0.01 for both tasks on a single GTX 1080Ti GPU.

\subsection{Correlations of Representations in Two Spaces}
\label{corr_two_space}

\textbf{Correlations with different CNN Models.} We firstly explored the correlation of temporal activations between the FBNs and convolutional filters in a variety of CNN models pre-trained on the ImageNet dataset \cite{deng2009imagenet} and Places365 dataset \cite{zhou2017places}. The filters in the last convolutional layer for all CNNs are selected and paired with FBNs. In Table~\ref{table1}, the averaged PCC values over all pairs across different CNN models and different runs of HCP 7T movie-watching dataset are reported. It is observed that for almost all CNN models across different runs (except the AlexNet, VGG-16, ResNet-18 at Run \#1 and/or Run \#2), the correlation measured by PCC is statistically significant with values larger than 0.2. The significance is well reproduced on the models pre-trained on different datasets (ImageNet and Places365) though the averaged PCC values may vary on different pre-trained models and runs.

\begin{table}[ht]
\centering
\setlength{\tabcolsep}{1.25mm}
\caption{The averaged PCC on HCP 7T movie-watching fMRI dataset for the pairs of FBNs and filters of the last convolutional layers on CNN models pre-trained on ImageNet and Places365 dataset. The correlations measured by PCC in this table are all statistically significant$(p$-value$ \leq 0.05)$ for different runs.}
\label{table1}
\begin{tabular}{lcccccccc}
\toprule

\multicolumn{1}{c}{\multirow{2}{*}{Methods}} & \multicolumn{4}{c}{ImageNet~\cite{deng2009imagenet}} & \multicolumn{4}{c}{Places365~\cite{zhou2017places}}\\
&  Run \#1 & Run \#2 & Run \#3 & Run \#4
&  Run \#1 & Run \#2 & Run \#3 & Run \#4\\ 
\midrule
AlexNet~\cite{krizhevsky2012imagenet} &0.2323 &0.2223 &0.2558 &0.2607 
&0.2651 &0.2374 &0.2788 &0.2774\\
VGG-16~\cite{simonyan2014very} &0.2376 &0.2176 &0.2654 &0.2617 & - & -&-&-\\
ResNet-18~\cite{he2016deep} &0.2415 &0.2267 &0.2516 &0.2530
&0.2703 &0.2536 &0.2896 &0.2931 \\
ResNet-50~\cite{he2016deep} &0.2862 &0.2660 &0.2942 &0.3022
&0.3008 &0.2745 &0.3076 &0.3159\\
DenseNet-161~\cite{huang2017densely} &0.3031 &0.2767 &0.3051 &0.3152 
&0.3052 &0.2879 &0.3134 &0.3199\\
Inception V3~\cite{szegedy2016rethinking} &0.2720 &0.2615 &0.2747 &0.2895& - & -&-&- \\
ShuffleNet V2~\cite{ma2018shufflenet} &0.2663 &0.2515 &0.2742 &0.2831 & - & -&-&- \\
MobileNet V2~\cite{sandler2018mobilenetv2} &0.2628 &0.2517 &0.2635 &0.2870& - & -&-&- \\
ResNeXt-50~\cite{xie2017aggregated} &0.2774 &0.2607 &0.2904 &0.2940 & - & -&-&- \\
MNASNet~\cite{tan2019mnasnet} &0.2612 &0.2422 &0.2669 &0.2699& - & -&-&- \\

\bottomrule
\end{tabular}
\end{table}

\begin{table}[ht]
\centering
\setlength{\tabcolsep}{1.25mm}
\caption{ 
The averaged PCC on StudyForrest movie-watching fMRI dataset for the pairs of FBNs and filters of the last convolutional layers on CNN models pre-trained on ImageNet dataset. The PCC value in this table with a * marker indicates that several pairs (less than 10) are not statistically significant$(p$-value$ \leq 0.05)$, otherwise it is significant.}
\label{table2}
\begin{tabular}{lccccccccc}
\toprule

\multicolumn{1}{c}{Methods}
&  Run \#1 & Run \#2 & Run \#3 & Run \#4
&  Run \#5 & Run \#6 & Run \#7 & Run \#8\\ 
\midrule
AlexNet~\cite{krizhevsky2012imagenet} &0.1369* &0.1659* &0.1712 &0.1540 &0.1609 &0.1724 &0.1590 &0.1961\\
VGG-16~\cite{simonyan2014very} &0.1398* &0.1836 &0.1822 &0.1777 &0.1752 &0.1839 &0.1732 &0.2112\\
ResNet-18~\cite{he2016deep} &0.1474* &0.1854 &0.1851 &0.1708 &0.1821 &0.1840 &0.1823 &0.2134 \\
ResNet-50~\cite{he2016deep} &0.1584 &0.2075 &0.2111 &0.1956
&0.1983 &0.2113 &0.1983 &0.2344\\
DenseNet-161~\cite{huang2017densely} &0.1673 &0.2143 &0.2154 &0.1985
&0.2049 &0.2158 &0.2049 &0.2497\\
Inception V3~\cite{szegedy2016rethinking} &0.1617 &0.2015 &0.2070 &0.1967 &0.1979 &0.2024 &0.1956 &0.2371 \\
ShuffleNet V2~\cite{ma2018shufflenet} &0.1488 &0.1922 &0.2004 &0.1900 &0.1898 &0.1986 &0.1920 &0.2324 \\
MobileNet V2~\cite{sandler2018mobilenetv2} &0.1499 &0.1961 &0.2001 &0.1898 &0.1906 &0.1958 &0.1929 &0.2350 \\
ResNeXt-50~\cite{xie2017aggregated} &0.1624 &0.2083 &0.2046 &0.1886 &0.1982 &0.2070 &0.1936 &0.2365 \\
MNASNet~\cite{tan2019mnasnet} &0.1553 &0.2036 &0.2015 &0.1907 &0.2013 &0.2029 &0.1974 &0.2387 \\

\bottomrule
\end{tabular}
\end{table}

We perform the similar analysis on the StudyForrest movie-watching fMRI dataset for validation, and the results are reported in Table~\ref{table2} with CNN models pre-trained on ImageNet dataset. However, the averaged PCC values are smaller than those on HCP 7T movie-watching dataset. This might be due to that the number of subjects in StudyForrest dataset (15) is smaller than those in HCP 7T fMRI dataset (184); the spatial/temporal resolution, image quality and signal-to-noise ratio of nfMRI data are worse than those in HCP 7T fMRI dataset. However, it is still found that the correlations are significant for almost all models and runs of fMRI except the AlexNet, VGG-16 and ResNet-18 on run \#1 and/or run \#2. Overall, these results consistently suggest that there exists significant correlation between the convolutional filters in CNN model and FBNs in human brain.

\textbf{Correlations with different CNN Layers.} We further assess the correlations of FBNs with convolutional filters in different layers of 4 different CNN models. The PCC values averaged over all FBN-filter pairs in each layer are reported in Table~\ref{table3}. We observe that the correlations are significant for pairs in the last two blocks/layers while some of them in the first two blocks/layers are not significant. The PCC values in different layers also show a trend that it increases and reaches a peak at the third layer, which is in line with the literature study \cite{yamins2014performance} reporting that the model’s intermediate layers are highly predictive of the brain's neural responses.

\begin{table}[ht]
\centering
\setlength{\tabcolsep}{1.25mm}
\caption{The averaged PCC for the pairs of FBNs and filters in different convolutional layers of CNN model and the ratio of \textbf{NOT} statistically significant pairs. The colors \textcolor{red}{red} and \textcolor{blue}{blue} denote the highest and the second-highest PCC value among different layers, respectively.}
\label{table3}
\begin{tabular}{lccccccccc}
\toprule

\multicolumn{1}{c}{\multirow{2}{*}{Methods}} & \multirow{2}{*}{Layer} & \multicolumn{2}{c}{Run 1} & \multicolumn{2}{c}{Run 2} & \multicolumn{2}{c}{Run 3} & \multicolumn{2}{c}{Run 4} \\
& & PCC & Ratio & PCC & Ratio & PCC & Ratio & PCC & Ratio  \\
\midrule
\multirow{4}{*}{ResNet-18~\cite{he2016deep}} 
 &Block \#1  &0.2401  &1/64  &0.2012  &2/64  &0.2189  &6/64  &0.2286  &0/64  \\
 &Block \#2  &\textcolor{blue}{0.2534}  &0/64  &0.2150  &0/64  &0.2513  &1/64  &0.2480  &0/64  \\
 &Block \#3  &\textcolor{red}{0.2743}  &0/64  &\textcolor{red}{0.2483}  &0/64  &\textcolor{red}{0.2821}  &0/64  &\textcolor{red}{0.2791}  &0/64  \\
 &Block \#4  &0.2415  &0/64  &\textcolor{blue}{0.2267}  &0/64  &\textcolor{blue}{0.2516}  &0/64  &\textcolor{blue}{0.2530}  &0/64  \\
\midrule
\multirow{4}{*}{ResNet-50~\cite{he2016deep}}
 &Block \#1  &0.2607  &0/64  &0.2276  &0/64  &0.2664  &0/64  &0.2504  &0/64  \\
 &Block \#2  &\textcolor{blue}{0.2876}  &0/64  &0.2498  &0/64  &0.2851  &0/64  &0.2768  &0/64  \\
 &Block \#3  &\textcolor{red}{0.2962}  &0/64  &\textcolor{red}{0.2684}  &0/64  &\textcolor{red}{0.3059}  &0/64  &\textcolor{red}{0.3060}  &0/64  \\
 &Block \#4  &0.2862  &0/64  &\textcolor{blue}{0.2660}  &0/64  &\textcolor{blue}{0.2942}  &0/64  &\textcolor{blue}{0.3022}  &0/64  \\
\midrule
\multirow{4}{*}{ShuffleNet V2~\cite{ma2018shufflenet}} 
 &Stage \#2  &0.2634  &0/64  &0.2226  &0/64  &0.2511  &1/64  &0.2460  &0/64  \\
 &Stage \#3  &0.2633  &0/64  &0.2365  &0/64  &0.2640  &0/64  &0.2647  &0/64  \\
 &Stage \#4  &\textcolor{red}{0.2749}  &0/64  &\textcolor{blue}{0.2511}  &0/64  &\textcolor{red}{0.2861}  &0/64  &\textcolor{red}{0.2968}  &0/64  \\
 &Conv \#5  &\textcolor{blue}{0.2663}  &0/64  &\textcolor{red}{0.2515}  &0/64  &\textcolor{blue}{0.2742}  &0/64  &\textcolor{blue}{0.2831}  &0/64  \\
\midrule
\multirow{4}{*}{ResNeXt-50~\cite{xie2017aggregated}} 
 &Block \#1  &0.2633  &0/64  &0.2280  &0/64  &0.2525  &2/64  &0.2478  &0/64  \\
 &Block \#2  &\textcolor{blue}{0.2803}  &0/64  &0.2501  &0/64  &0.2840  &0/64  &0.2769  &0/64  \\
 &Block \#3  &\textcolor{red}{0.2954}  &0/64  &\textcolor{red}{0.2635}  &0/64  &\textcolor{red}{0.3026}  &0/64  &\textcolor{red}{0.3030}  &0/64  \\
 &Block \#4  &0.2774  &0/64  &\textcolor{blue}{0.2607}  &0/64  &\textcolor{blue}{0.2904}  &0/64  &\textcolor{blue}{0.2940}  &0/64  \\
\bottomrule
\end{tabular}
\end{table}

\begin{figure}[ht]
  \centering
  \includegraphics[width=0.9\linewidth]{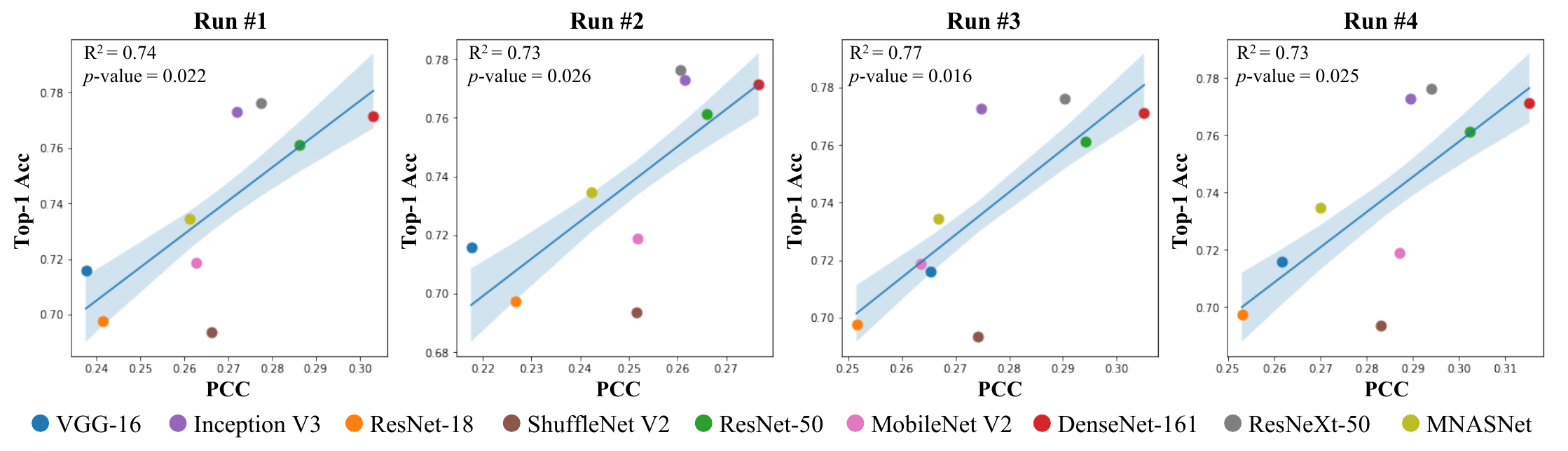}
  \caption{The linear regression modeling the relationship between PCC and CNN's top-1 image classification accuracy on ImageNet. Different CNN models are marked as circle with different color.}
  \label{figure3}
\end{figure}

\begin{figure}[ht]
  \centering
  \includegraphics[width=0.9\linewidth]{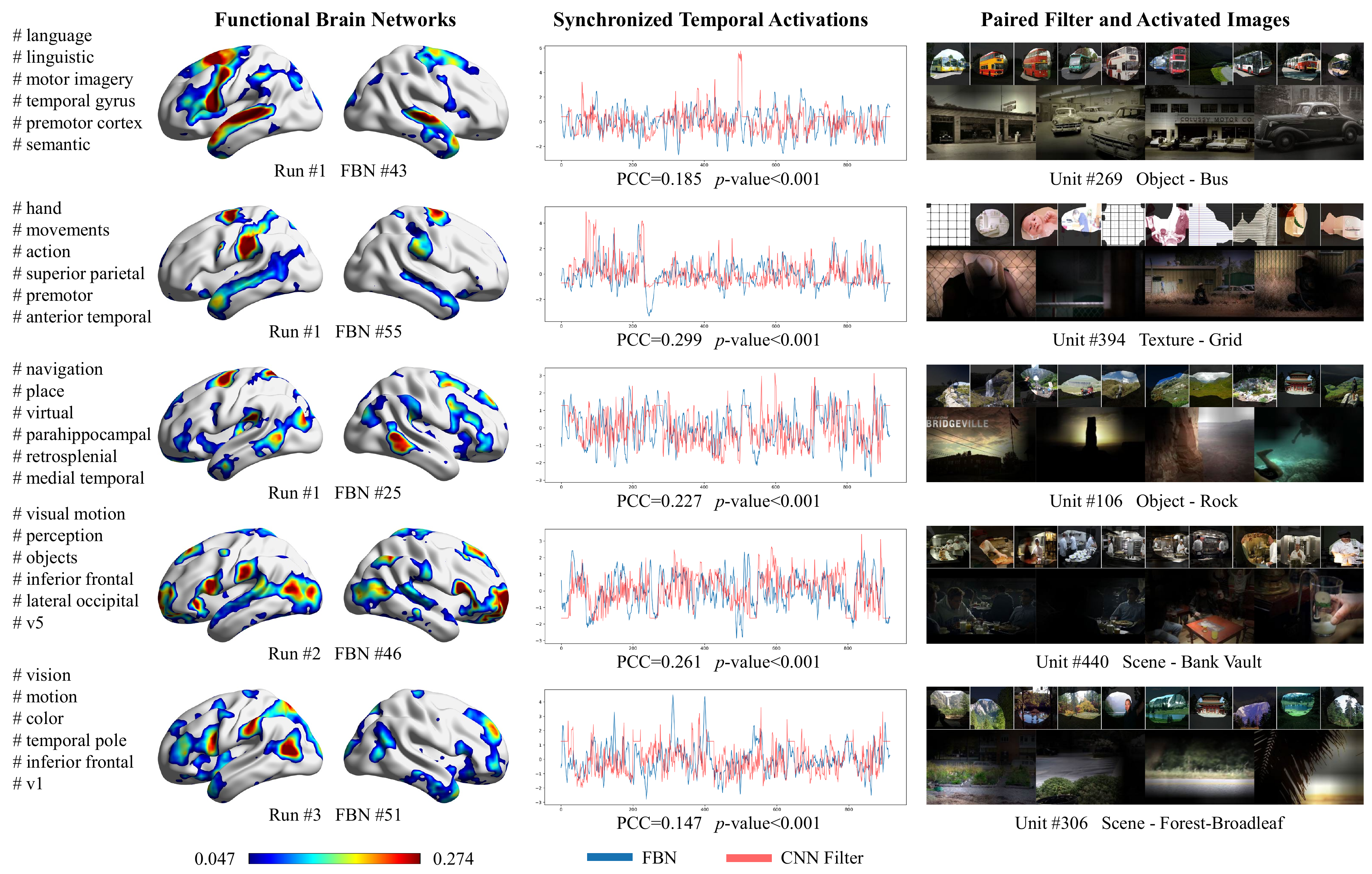}
  \caption{The visualization of FBN-Filter pairs obtained from our model. The left panel is the FBNs to be paired and semantic description from fMRI meta-analysis. The middle panel shows the synchronized activations from FBN and paired CNN filter. The right panel shows the most activated frames and the corresponding semantic description and filter's representative images in ~\cite{bau2017network}.}
  \label{figure4}
\end{figure}

\subsection{PCC Variance in Different CNN Models} 
From Table~\ref{table1} and Table~\ref{table2}, it is observed that the ResNet-50 and DenseNet-161 have higher PCC values than the VGG-16 and ResNet-18, which suggests that the PCC values may have correlations with the model's representation ability and performance. To verify this hypothesis, we conduct the linear regression to model the relationship between the PCC values and the CNN model's top-1 accuracy on ImageNet classification task. The results are reported in 
Fig.~\ref{figure3}. We found that the relationship can be represented well by the linear model with $R^2$ larger than 0.7 and $p$-value smaller than 0.05 across all 4 runs of HCP 7T movie-watching task fMRI dataset. This result indicates that if the visual representations of CNN models are more similar to those of human brain function, its performance on image classification task will be better. This is consistent with the study~\cite{you2020graph} finding that the top-performing ANNs have a graph structure similar to those of BNNs.

\subsection{Visualizations of the Cross-Annotation}
We conduct the cross-annotation based on the Eq.~\eqref{eq2} to pair each FBN with a filter at the last convolutional layer of ResNet-18 and visualize several sample pairs in Fig.~\ref{figure4}. The left panel in Fig.~\ref{figure4} shows the FBNs to be paired and the corresponding semantic description from fMRI meta-analysis. The right panel shows the most activated images obtained by \cite{bau2017network} from the movie frame sequence of paired CNN filters. The filter's corresponding semantic description and representative images are also demonstrated. In Fig.~\ref{figure4}, we found some interesting connections about the semantic description of the paired FBN and filters. For example, the description of FBN \#25 is related to place and navigation, while the paired filters are labeled as rock and the representative images are related to some natural scenes. Such observation is obvious on some pairs, which is congruent with the results in Section~\ref{corr_two_space}. We provided more samples with other CNN models as well as the cross-annotation based on Eq.~\eqref{eq1} in supplementary materials for comparisons.

\subsection{Ablation Studies of fMRI Embedding Framework}
We conduct the ablation studies for our fMRI embedding framework on three variants: a) encoder/decoder only has one linear transformation (LT) layer; b) encoder/decoder has one LT layer and two LSTM layers with tanh activation function; c) encoder/decoder has one LT layer followed by multi-head self-attention module. We measure the similarity of temporal activations of FBN-filter pairs identified by PCC. The averaged values for different metrics over all pairs and runs are reported in Table~\ref{table4}. Overall, the similarity of LT+LSTM and LT+MSA is larger than the linear transformation baseline. The LT+LSTM and LT+MSA have comparable performances in terms of similarity. However, the LT+MSA have better FBNs quality. We provide the details in the supplementary materials. The results and analysis of our work are based on LT+MSA.

\begin{table}
\centering
\setlength{\tabcolsep}{1.25mm}
\caption{The comparison of temporal activations similarity of FBN-filter pairs identified by PCC. The colors \textcolor{red}{red} and \textcolor{blue}{blue} denote the best and the second-best results, respectively. Abbreviations: LT: linear transformation; MSA: multi-head self-attention.}
\label{table4}
\begin{tabular}{lcccccccc}
\toprule

Methods &  MAE$\downarrow$  & MSE$\downarrow$  & RMSE$\downarrow$   & DTW$\downarrow$ & PCC$\uparrow$\\ 
\midrule

a) LT &0.9781 &1.5309  &1.2360  & \textcolor{red}{18.0688} &0.2346\\
b) LT+LSTM &\textcolor{red}{0.9392} &\textcolor{red}{1.4411}  &\textcolor{red}{1.1986}  & \textcolor{blue}{18.4683} &\textcolor{red}{0.2794} \\
c) LT+MSA &\textcolor{blue}{0.9658} &\textcolor{blue}{1.5136}  &\textcolor{blue}{1.2289}  & 18.7328 &\textcolor{blue}{0.2432} \\
\bottomrule
\end{tabular}
\end{table}

\section{Discussions}
\textbf{Interpretability}. The proposed Sync-ACT framework matches and pairs the neurons in ANNs and BNNs, based on which the cross-annotation is performed to annotate the neurons in one domain with the semantic description in the other. This scheme allows us to study and interpret the ANN and its behavior from a new perspective: how it is similar to the function of human brain. The Sync-ACT framework opens a new paradigm for the interpretability studies of ANN by using the prior knowledge in neuroscience to interpret ANNs. In parallel, the FBNs are complex and dynamic under naturalistic stimuli and it is difficult to characterize their roles and functions. With Sync-ACT, we can understand the dynamic function of FBNs with visual and language description from the paired ANNs' neurons in a direct way, providing a novel way for unveiling the complex brain function.

\textbf{Neural architecture search}. One important finding of this study is that the performance of CNNs on image classification task is closely related to its visual representation similarity with the human brain. In the literature~\cite{elsken2019neural,liu2018darts,zoph2016neural}, the typical evaluation criteria for neural architecture search (NAS) is the performance of searched neural network on a specific task. Our Sync-ACT framework provides new inspirations: the ANN's representation similarity to human brain could be a viable and meaningful criteria for NAS and thus guide the NAS approaches to increase ANNs' similarities with BNNs in human brain in order to improve the interpretability and performance. Overall, our Sync-ACT framework contributes to the emerging field of brain-inspired AI, i.e., using domain knowledge of brain science to inspire and guide the design of AI models and their applications.

\textbf{Limitations.} Our approach has several potential limitations. a) We use the maximum value in the feature map to represent the activation degree of convolutional filters. Currently, how to characterize the activation degree of filters is still an open question, which deserves more efforts in the future. b) The semantics descriptions from meta-analysis~\cite{yarkoni2011large} and Bau et al.~\cite{bau2017network} for neurons in FBNs and CNNs are coarse-grained and ad-hoc (e.g., unit \#440, Bank Vault) due to their intrinsic limitations. More fine-grained description can be explored and adopted in the future. c) We mainly focus on the CNNs for image classification. CNNs and ViTs for other tasks should be investigated in the future.

\section{Conclusion}
In this paper, we proposed a novel computational framework, Sync-ACT, to couple the visual representation spaces and semantics between ANNs and BNNs in human brain by synchronizing their activations to visual stimulus. With this approach, we found the significant correlation of the semantics between the visual representations in CNNs and those in human brain. Also, CNN's visual representation similarity to human brain is closely related to its performance on image classification task. In the future, our Sync-ACT model can be easily extended to other naturalistic stimuli such as natural language and/or audio to explore the connection of natural language processing (NLP) model's semantic space with the one in human brain. Overall, our study introduces a general and effective paradigm to couple the ANNs and BNNs and provides novel insight on their connections.

\bibliographystyle{splncs04}

\begin{thebibliography}{10}
\providecommand{\url}[1]{\texttt{#1}}
\providecommand{\urlprefix}{URL }
\providecommand{\doi}[1]{https://doi.org/#1}

\bibitem{barch2013function}
Barch, D.M., Burgess, G.C., Harms, M.P., Petersen, S.E., Schlaggar, B.L.,
  Corbetta, M., Glasser, M.F., Curtiss, S., Dixit, S., Feldt, C., et~al.:
  Function in the human connectome: task-fmri and individual differences in
  behavior. Neuroimage  \textbf{80},  169--189 (2013)

\bibitem{barrett2019analyzing}
Barrett, D.G., Morcos, A.S., Macke, J.H.: Analyzing biological and artificial
  neural networks: challenges with opportunities for synergy? Current opinion
  in neurobiology  \textbf{55},  55--64 (2019)

\bibitem{bau2017network}
Bau, D., Zhou, B., Khosla, A., Oliva, A., Torralba, A.: Network dissection:
  Quantifying interpretability of deep visual representations. In: Proceedings
  of the IEEE conference on computer vision and pattern recognition. pp.
  6541--6549 (2017)

\bibitem{calhoun2006unmixing}
Calhoun, V.D., Adali, T.: Unmixing fmri with independent component analysis.
  IEEE Engineering in Medicine and Biology Magazine  \textbf{25}(2),  79--90
  (2006)

\bibitem{dalvi2019one}
Dalvi, F., Durrani, N., Sajjad, H., Belinkov, Y., Bau, A., Glass, J.: What is
  one grain of sand in the desert? analyzing individual neurons in deep nlp
  models. In: Proceedings of the AAAI Conference on Artificial Intelligence.
  vol.~33, pp. 6309--6317 (2019)

\bibitem{deng2009imagenet}
Deng, J., Dong, W., Socher, R., Li, L.J., Li, K., Fei-Fei, L.: Imagenet: A
  large-scale hierarchical image database. In: 2009 IEEE conference on computer
  vision and pattern recognition. pp. 248--255. Ieee (2009)

\bibitem{dong2019modeling}
Dong, Q., Ge, F., Ning, Q., Zhao, Y., Lv, J., Huang, H., Yuan, J., Jiang, X.,
  Shen, D., Liu, T.: Modeling hierarchical brain networks via volumetric sparse
  deep belief network. IEEE transactions on biomedical engineering
  \textbf{67}(6),  1739--1748 (2019)

\bibitem{dosovitskiy2020image}
Dosovitskiy, A., Beyer, L., Kolesnikov, A., Weissenborn, D., Zhai, X.,
  Unterthiner, T., Dehghani, M., Minderer, M., Heigold, G., Gelly, S., et~al.:
  An image is worth 16x16 words: Transformers for image recognition at scale.
  arXiv preprint arXiv:2010.11929  (2020)

\bibitem{elsken2019neural}
Elsken, T., Metzen, J.H., Hutter, F.: Neural architecture search: A survey. The
  Journal of Machine Learning Research  \textbf{20}(1),  1997--2017 (2019)

\bibitem{esteban2019fmriprep}
Esteban, O., Markiewicz, C.J., Blair, R.W., Moodie, C.A., Isik, A.I.,
  Erramuzpe, A., Kent, J.D., Goncalves, M., DuPre, E., Snyder, M., et~al.:
  fmriprep: a robust preprocessing pipeline for functional mri. Nature methods
  \textbf{16}(1),  111--116 (2019)

\bibitem{glasser2013minimal}
Glasser, M.F., Sotiropoulos, S.N., Wilson, J.A., Coalson, T.S., Fischl, B.,
  Andersson, J.L., Xu, J., Jbabdi, S., Webster, M., Polimeni, J.R., et~al.: The
  minimal preprocessing pipelines for the human connectome project. Neuroimage
  \textbf{80},  105--124 (2013)

\bibitem{golland2007extrinsic}
Golland, Y., Bentin, S., Gelbard, H., Benjamini, Y., Heller, R., Nir, Y.,
  Hasson, U., Malach, R.: Extrinsic and intrinsic systems in the posterior
  cortex of the human brain revealed during natural sensory stimulation.
  Cerebral cortex  \textbf{17}(4),  766--777 (2007)

\bibitem{hanke2016studyforrest}
Hanke, M., Adelh{\"o}fer, N., Kottke, D., Iacovella, V., Sengupta, A., Kaule,
  F.R., Nigbur, R., Waite, A.Q., Baumgartner, F., Stadler, J.: A studyforrest
  extension, simultaneous fmri and eye gaze recordings during prolonged natural
  stimulation. Scientific data  \textbf{3}(1),  1--15 (2016)

\bibitem{he2016deep}
He, K., Zhang, X., Ren, S., Sun, J.: Deep residual learning for image
  recognition. In: Proceedings of the IEEE conference on computer vision and
  pattern recognition. pp. 770--778 (2016)

\bibitem{hernandez2021natural}
Hernandez, E., Schwettmann, S., Bau, D., Bagashvili, T., Torralba, A., Andreas,
  J.: Natural language descriptions of deep features. In: International
  Conference on Learning Representations (2021)

\bibitem{hochreiter1997long}
Hochreiter, S., Schmidhuber, J.: Long short-term memory. Neural computation
  \textbf{9}(8),  1735--1780 (1997)

\bibitem{hu2010bridging}
Hu, X., Deng, F., Li, K., Zhang, T., Chen, H., Jiang, X., Lv, J., Zhu, D.,
  Faraco, C., Zhang, D., et~al.: Bridging low-level features and high-level
  semantics via fmri brain imaging for video classification. In: Proceedings of
  the 18th ACM international conference on Multimedia. pp. 451--460 (2010)

\bibitem{huang2017densely}
Huang, G., Liu, Z., Van Der~Maaten, L., Weinberger, K.Q.: Densely connected
  convolutional networks. In: Proceedings of the IEEE conference on computer
  vision and pattern recognition. pp. 4700--4708 (2017)

\bibitem{khan2020survey}
Khan, A., Sohail, A., Zahoora, U., Qureshi, A.S.: A survey of the recent
  architectures of deep convolutional neural networks. Artificial intelligence
  review  \textbf{53}(8),  5455--5516 (2020)

\bibitem{kim2016convolutional}
Kim, J., Sangjun, O., Kim, Y., Lee, M.: Convolutional neural network with
  biologically inspired retinal structure. Procedia Computer Science
  \textbf{88},  145--154 (2016)

\bibitem{kingma2014adam}
Kingma, D.P., Ba, J.: Adam: A method for stochastic optimization. arXiv
  preprint arXiv:1412.6980  (2014)

\bibitem{krizhevsky2012imagenet}
Krizhevsky, A., Sutskever, I., Hinton, G.E.: Imagenet classification with deep
  convolutional neural networks. Advances in neural information processing
  systems  \textbf{25} (2012)

\bibitem{lecun2015deep}
LeCun, Y., Bengio, Y., Hinton, G.: Deep learning. nature  \textbf{521}(7553),
  436--444 (2015)

\bibitem{lecun1995convolutional}
LeCun, Y., Bengio, Y., et~al.: Convolutional networks for images, speech, and
  time series. The handbook of brain theory and neural networks
  \textbf{3361}(10), ~1995 (1995)

\bibitem{li2021simultaneous}
Li, Q., Dong, Q., Ge, F., Qiang, N., Wu, X., Liu, T.: Simultaneous
  spatial-temporal decomposition for connectome-scale brain networks by deep
  sparse recurrent auto-encoder. Brain Imaging and Behavior  \textbf{15}(5),
  2646--2660 (2021)

\bibitem{li2021evolutional}
Li, Q., Zhang, W., Zhao, L., Wu, X., Liu, T.: Evolutional neural architecture
  search for optimization of spatiotemporal brain network decomposition. IEEE
  Transactions on Biomedical Engineering  (2021)

\bibitem{liu2018darts}
Liu, H., Simonyan, K., Yang, Y.: Darts: Differentiable architecture search.
  arXiv preprint arXiv:1806.09055  (2018)

\bibitem{liu2019cerebral}
Liu, H., Zhang, S., Jiang, X., Zhang, T., Huang, H., Ge, F., Zhao, L., Li, X.,
  Hu, X., Han, J., et~al.: The cerebral cortex is bisectionally segregated into
  two fundamentally different functional units of gyri and sulci. Cerebral
  Cortex  \textbf{29}(10),  4238--4252 (2019)

\bibitem{liu2014merging}
Liu, T., Hu, X., Li, X., Chen, M., Han, J., Guo, L.: Merging neuroimaging and
  multimedia: Methods, opportunities, and challenges. IEEE Transactions on
  Human-Machine Systems  \textbf{44}(2),  270--280 (2014)

\bibitem{luo2016understanding}
Luo, W., Li, Y., Urtasun, R., Zemel, R.: Understanding the effective receptive
  field in deep convolutional neural networks. Advances in neural information
  processing systems  \textbf{29} (2016)

\bibitem{lv2014holistic}
Lv, J., Jiang, X., Li, X., Zhu, D., Zhang, S., Zhao, S., Chen, H., Zhang, T.,
  Hu, X., Han, J., et~al.: Holistic atlases of functional networks and
  interactions reveal reciprocal organizational architecture of cortical
  function. IEEE Transactions on Biomedical Engineering  \textbf{62}(4),
  1120--1131 (2014)

\bibitem{ma2018shufflenet}
Ma, N., Zhang, X., Zheng, H.T., Sun, J.: Shufflenet v2: Practical guidelines
  for efficient cnn architecture design. In: Proceedings of the European
  conference on computer vision (ECCV). pp. 116--131 (2018)

\bibitem{mahendran2016visualizing}
Mahendran, A., Vedaldi, A.: Visualizing deep convolutional neural networks
  using natural pre-images. International Journal of Computer Vision
  \textbf{120}(3),  233--255 (2016)

\bibitem{morcos2018importance}
Morcos, A.S., Barrett, D.G., Rabinowitz, N.C., Botvinick, M.: On the importance
  of single directions for generalization. arXiv preprint arXiv:1803.06959
  (2018)

\bibitem{mu2020compositional}
Mu, J., Andreas, J.: Compositional explanations of neurons. Advances in Neural
  Information Processing Systems  \textbf{33},  17153--17163 (2020)

\bibitem{mwangi2014review}
Mwangi, B., Tian, T.S., Soares, J.C.: A review of feature reduction techniques
  in neuroimaging. Neuroinformatics  \textbf{12}(2),  229--244 (2014)

\bibitem{ren2017inter}
Ren, Y., Nguyen, V.T., Guo, L., Guo, C.C.: Inter-subject functional correlation
  reveal a hierarchical organization of extrinsic and intrinsic systems in the
  brain. Scientific reports  \textbf{7}(1),  1--12 (2017)

\bibitem{ren2021hierarchical}
Ren, Y., Xu, S., Tao, Z., Song, L., He, X.: Hierarchical spatio-temporal
  modeling of naturalistic functional magnetic resonance imaging signals via
  two-stage deep belief network with neural architecture search. Frontiers in
  Neuroscience  \textbf{15},  794955 (2021)

\bibitem{sandler2018mobilenetv2}
Sandler, M., Howard, A., Zhu, M., Zhmoginov, A., Chen, L.C.: Mobilenetv2:
  Inverted residuals and linear bottlenecks. In: Proceedings of the IEEE
  conference on computer vision and pattern recognition. pp. 4510--4520 (2018)

\bibitem{simonyan2014very}
Simonyan, K., Zisserman, A.: Very deep convolutional networks for large-scale
  image recognition. arXiv preprint arXiv:1409.1556  (2014)

\bibitem{szegedy2016rethinking}
Szegedy, C., Vanhoucke, V., Ioffe, S., Shlens, J., Wojna, Z.: Rethinking the
  inception architecture for computer vision. In: Proceedings of the IEEE
  conference on computer vision and pattern recognition. pp. 2818--2826 (2016)

\bibitem{tan2019mnasnet}
Tan, M., Chen, B., Pang, R., Vasudevan, V., Sandler, M., Howard, A., Le, Q.V.:
  Mnasnet: Platform-aware neural architecture search for mobile. In:
  Proceedings of the IEEE/CVF Conference on Computer Vision and Pattern
  Recognition. pp. 2820--2828 (2019)

\bibitem{van2013wu}
Van~Essen, D.C., Smith, S.M., Barch, D.M., Behrens, T.E., Yacoub, E., Ugurbil,
  K., Consortium, W.M.H., et~al.: The wu-minn human connectome project: an
  overview. Neuroimage  \textbf{80},  62--79 (2013)

\bibitem{vaswani2017attention}
Vaswani, A., Shazeer, N., Parmar, N., Uszkoreit, J., Jones, L., Gomez, A.N.,
  Kaiser, {\L}., Polosukhin, I.: Attention is all you need. Advances in neural
  information processing systems  \textbf{30} (2017)

\bibitem{wang2018recognizing}
Wang, H., Zhao, S., Dong, Q., Cui, Y., Chen, Y., Han, J., Xie, L., Liu, T.:
  Recognizing brain states using deep sparse recurrent neural network. IEEE
  transactions on medical imaging  \textbf{38}(4),  1058--1068 (2018)

\bibitem{wang2017test}
Wang, J., Ren, Y., Hu, X., Nguyen, V.T., Guo, L., Han, J., Guo, C.C.:
  Test--retest reliability of functional connectivity networks during
  naturalistic fmri paradigms. Human brain mapping  \textbf{38}(4),  2226--2241
  (2017)

\bibitem{xie2017aggregated}
Xie, S., Girshick, R., Doll{\'a}r, P., Tu, Z., He, K.: Aggregated residual
  transformations for deep neural networks. In: Proceedings of the IEEE
  conference on computer vision and pattern recognition. pp. 1492--1500 (2017)

\bibitem{yamins2016using}
Yamins, D.L., DiCarlo, J.J.: Using goal-driven deep learning models to
  understand sensory cortex. Nature neuroscience  \textbf{19}(3),  356--365
  (2016)

\bibitem{yamins2014performance}
Yamins, D.L., Hong, H., Cadieu, C.F., Solomon, E.A., Seibert, D., DiCarlo,
  J.J.: Performance-optimized hierarchical models predict neural responses in
  higher visual cortex. Proceedings of the national academy of sciences
  \textbf{111}(23),  8619--8624 (2014)

\bibitem{yarkoni2011large}
Yarkoni, T., Poldrack, R.A., Nichols, T.E., Van~Essen, D.C., Wager, T.D.:
  Large-scale automated synthesis of human functional neuroimaging data. Nature
  methods  \textbf{8}(8),  665--670 (2011)

\bibitem{you2020graph}
You, J., Leskovec, J., He, K., Xie, S.: Graph structure of neural networks. In:
  International Conference on Machine Learning. pp. 10881--10891. PMLR (2020)

\bibitem{zhang2019identify}
Zhang, W., Zhao, L., Li, Q., Zhao, S., Dong, Q., Jiang, X., Zhang, T., Liu, T.:
  Identify hierarchical structures from task-based fmri data via hybrid
  spatiotemporal neural architecture search net. In: International Conference
  on Medical Image Computing and Computer-Assisted Intervention. pp. 745--753.
  Springer (2019)

\bibitem{zhao2021exploring}
Zhao, L., Dai, H., Jiang, X., Zhang, T., Zhu, D., Liu, T.: Exploring the
  functional difference of gyri/sulci via hierarchical interpretable
  autoencoder. In: International Conference on Medical Image Computing and
  Computer-Assisted Intervention. pp. 701--709. Springer (2021)

\bibitem{zhou2016learning}
Zhou, B., Khosla, A., Lapedriza, A., Oliva, A., Torralba, A.: Learning deep
  features for discriminative localization. In: Proceedings of the IEEE
  conference on computer vision and pattern recognition. pp. 2921--2929 (2016)

\bibitem{zhou2017places}
Zhou, B., Lapedriza, A., Khosla, A., Oliva, A., Torralba, A.: Places: A 10
  million image database for scene recognition. IEEE transactions on pattern
  analysis and machine intelligence  \textbf{40}(6),  1452--1464 (2017)

\bibitem{zoph2016neural}
Zoph, B., Le, Q.V.: Neural architecture search with reinforcement learning.
  arXiv preprint arXiv:1611.01578  (2016)

\end{thebibliography}

\newpage

\appendix

\section{Datasets and Preprocessing}
We provide the dataset description and preprocessing details in this section. 

For HCP 7T movie-watching fMRI dataset (\url{http://www.humanconnectomeproject.org/}) of S1200 release \cite{barch2013function}, the important acquisition parameters are as follows: 130$\times$130 matrix, 85 slices, TR=1.0 $s$, TE=22.2 $s$, 208 $mm$ FOV, flip angle = 45$^{\circ}$, 1.6 $mm$ isotropic voxels. The movie clips have a resolution of 1024$\times$720 pixels (24fps). We extracted the last movie frame in each second as the corresponding image for fMRI data and resized it with a resolution of 256$\times$180 pixels. 

For StudyForrest movie-wathcing fMRI dataset (\url{https://www.studyforrest.org/}) \cite{hanke2016studyforrest}, the important acquisition parameters are as follows: 80$\times$80 matrix, 35 slices, TR=2.0 $s$, TE=30.0 $ms$, 240 $mm$ FOV, flip angle = 90$^{\circ}$, 3.0 $mm$ isotropic voxels.  The movie clips have a resolution of 1280$\times$720 pixels (25fps). We extracted the last movie frame every two seconds as the corresponding image for fMRI data and resized it with a resolution of 320$\times$180 pixels.

\section{Influence of Number of FBNs}
In this section, we investigate the influence of the number of FBNs. It is noted that there is no golden rule for the number of FBNs to be derived from the model \cite{lv2014holistic}. So, we empirically set the number of FBNs to 64/128/256/512 for investigating its influence. We reported the correlation of temporal activations between the FBNs and convolutional filters in the Table \ref{table1}, Table \ref{table2} and Table \ref{table3}. The filters in the last convolutional layer for all CNNs are selected and paired with FBNs. 

\begin{table}[h]
\centering
\setlength{\tabcolsep}{1.25mm}
\caption{The averaged PCC on HCP 7T movie-watching fMRI dataset for the pairs of \textbf{128} FBNs and filters of the last convolutional layers on CNN models pre-trained on ImageNet and Places365 dataset. The correlations measured by PCC in this table are all statistically significant$(p$-value$ \leq 0.05)$ for different runs.}
\label{table1}
\begin{tabular}{lcccccccc}
\toprule

\multicolumn{1}{c}{\multirow{2}{*}{Methods}} & \multicolumn{4}{c}{ImageNet~\cite{deng2009imagenet}} & \multicolumn{4}{c}{Places365~\cite{zhou2017places}}\\
&  Run \#1 & Run \#2 & Run \#3 & Run \#4
&  Run \#1 & Run \#2 & Run \#3 & Run \#4\\ 
\midrule
AlexNet~\cite{krizhevsky2012imagenet} &0.2422 &0.2324 &0.2650 &0.2532 
&0.2693 &0.2489 &0.2913 &0.2713\\
VGG-16~\cite{simonyan2014very} &0.2514 &0.2272 &0.2741 &0.2680 & - & -&-&-\\
ResNet-18~\cite{he2016deep} &0.2507 &0.2382 &0.2607 &0.2511
&0.2792 &0.2603 &0.3015 &0.2875 \\
ResNet-50~\cite{he2016deep} &0.2927 &0.2750 &0.3029 &0.3005
&0.3088 &0.2905 &0.3222 &0.3148\\
DenseNet-161~\cite{huang2017densely} &0.3067 &0.2886 &0.3146 &0.3132 
&0.3137 &0.2949 &0.3266 &0.3200\\
Inception V3~\cite{szegedy2016rethinking} &0.2735 &0.2703 &0.2823 &0.2848 & - & -&-&- \\
ShuffleNet V2~\cite{ma2018shufflenet} &0.2724 &0.2632 &0.2874 &0.2797 & - & -&-&- \\
MobileNet V2~\cite{sandler2018mobilenetv2} &0.2718 &0.2634 &0.2830 &0.2901 & - & -&-&- \\
ResNeXt-50~\cite{xie2017aggregated} &0.2810 &0.2763 &0.2983 &0.2943 & - & -&-&- \\
MNASNet~\cite{tan2019mnasnet} &0.2667 &0.2535 &0.2765 &0.2722 & - & -&-&- \\

\bottomrule
\end{tabular}
\end{table}

\begin{table}[h]
\centering
\setlength{\tabcolsep}{1.25mm}
\caption{The averaged PCC on HCP 7T movie-watching fMRI dataset for the pairs of \textbf{256} FBNs and filters of the last convolutional layers on CNN models pre-trained on ImageNet and Places365 dataset. The correlations measured by PCC in this table are all statistically significant$(p$-value$ \leq 0.05)$ for different runs.}
\label{table2}
\begin{tabular}{lcccccccc}
\toprule

\multicolumn{1}{c}{\multirow{2}{*}{Methods}} & \multicolumn{4}{c}{ImageNet~\cite{deng2009imagenet}} & \multicolumn{4}{c}{Places365~\cite{zhou2017places}}\\
&  Run \#1 & Run \#2 & Run \#3 & Run \#4
&  Run \#1 & Run \#2 & Run \#3 & Run \#4\\ 
\midrule
AlexNet~\cite{krizhevsky2012imagenet} &0.2612 &0.2601 &0.3266 &0.2850 
&0.3041 &0.2806 &0.3633 &0.3160\\
VGG-16~\cite{simonyan2014very} &0.2711 &0.2484 &0.3319 &0.3007 & - & -&-&-\\
ResNet-18~\cite{he2016deep} &0.2763 &0.2698 &0.2992 &0.2861
&0.3049 &0.2887 &0.3595 &0.3200 \\
ResNet-50~\cite{he2016deep} &0.3250 &0.3102 &0.3740 &0.3437
&0.3482 &0.3264 &0.3899 &0.3509\\
DenseNet-161~\cite{huang2017densely} &0.3400 &0.3278 &0.3975 &0.3608 
&0.3473 &0.3379 &0.4053 &0.3707\\
Inception V3~\cite{szegedy2016rethinking} &0.3071 &0.2946 &0.3506 &0.3238 & - & -&-&- \\
ShuffleNet V2~\cite{ma2018shufflenet} &0.3084 &0.2956 &0.3581 &0.3205 & - & -&-&- \\
MobileNet V2~\cite{sandler2018mobilenetv2} &0.3071 &0.2969 &0.3398 &0.3334 & - & -&-&- \\
ResNeXt-50~\cite{xie2017aggregated} &0.3073 &0.3103 &0.3696 &0.3456 & - & -&-&- \\
MNASNet~\cite{tan2019mnasnet} &0.2961 &0.2873 &0.3241 &0.3037 & - & -&-&- \\

\bottomrule
\end{tabular}
\end{table}

\begin{table}[h]
\centering
\setlength{\tabcolsep}{1.25mm}
\caption{The averaged PCC on HCP 7T movie-watching fMRI dataset for the pairs of \textbf{512} FBNs and filters of the last convolutional layers on CNN models pre-trained on ImageNet and Places365 dataset. The correlations measured by PCC in this table are all statistically significant$(p$-value$ \leq 0.05)$ for different runs.}
\label{table3}
\begin{tabular}{lcccccccc}
\toprule

\multicolumn{1}{c}{\multirow{2}{*}{Methods}} & \multicolumn{4}{c}{ImageNet~\cite{deng2009imagenet}} & \multicolumn{4}{c}{Places365~\cite{zhou2017places}}\\
&  Run \#1 & Run \#2 & Run \#3 & Run \#4
&  Run \#1 & Run \#2 & Run \#3 & Run \#4\\ 
\midrule
AlexNet~\cite{krizhevsky2012imagenet} &0.2485 &0.2667 &0.3011 &0.2925 
&0.2863 &0.2854 &0.3338 &0.3194\\
VGG-16~\cite{simonyan2014very} &0.2520 &0.2508 &0.3009 &0.3017 & - & -&-&-\\
ResNet-18~\cite{he2016deep} &0.2674 &0.2724 &0.2834 &0.2815
&0.2941 &0.3012 &0.3238 &0.3224 \\
ResNet-50~\cite{he2016deep} &0.3076 &0.3142 &0.3390 &0.3441
&0.3335 &0.3308 &0.3526 &0.3570\\
DenseNet-161~\cite{huang2017densely} &0.3203 &0.3331 &0.3619 &0.3649 
&0.3282 &0.3446 &0.3685 &0.3708\\
Inception V3~\cite{szegedy2016rethinking} &0.2883 &0.3009 &0.3190 &0.3243 & - & -&-&- \\
ShuffleNet V2~\cite{ma2018shufflenet} &0.2875 &0.2991 &0.3216 &0.3254 & - & -&-&- \\
MobileNet V2~\cite{sandler2018mobilenetv2} &0.2874 &0.3082 &0.3075 &0.3304 & - & -&-&- \\
ResNeXt-50~\cite{xie2017aggregated} &0.2946 &0.3158 &0.3302 &0.3438 & - & -&-&- \\
MNASNet~\cite{tan2019mnasnet} &0.2755 &0.2895 &0.2985 &0.3046 & - & -&-&- \\

\bottomrule
\end{tabular}
\end{table}

It is observed that for all CNN models, the correlation measured by PCC is statistically significant with values larger than 0.2 for 128/256/512 FBNs, which is consistent with the Table 1 in the main manuscript for 64 FBNs. We also found that the averaged PCC increases from 64 FBNs and reaches the peak with 256 FBNs, while decreases with 512 FBNs. Overall, these results suggest that the observation with 64 FBNs can be well reproduced with different number of FBNs and is not obtained by chance. 

\section{Visualizations of Cross-Annotation at Different Layers}
In this section, we provide more examples for the cross-annotation based on the Eq. (2) to pair each FBN with a filter at the $1^{st}$/$2^{nd}$/$3^{rd}$ block of ResNet-18 and visualize several sample pairs in Fig.~\ref{figure1}, Fig.~\ref{figure2} and Fig.~\ref{figure3}, respectively.

\begin{figure}[h!]
  \centering
  \includegraphics[width=0.9\linewidth]{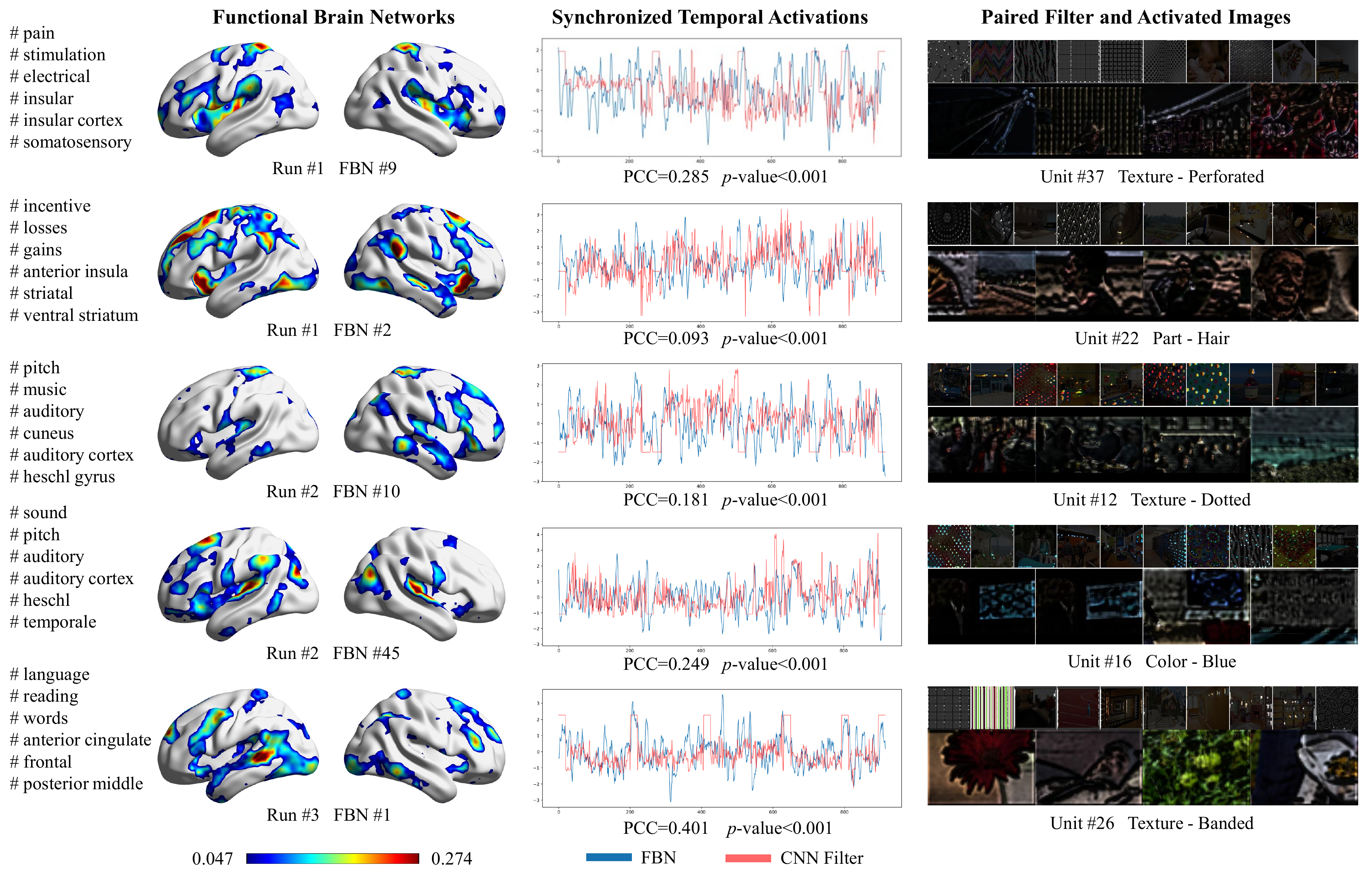}
  \caption{The visualization of FBN-Filter pairs obtained from ResNet-18 Block \#1. The left panel is the FBNs to be paired and semantic description from fMRI meta-analysis. The middle panel shows the synchronized activations from FBN and paired CNN filter. The right panel shows the most activated frames and the corresponding semantic description and filter's representative images~\cite{bau2017network}.}
  \label{figure1}
\end{figure}

\begin{figure}[h!]
  \centering
  \includegraphics[width=0.9\linewidth]{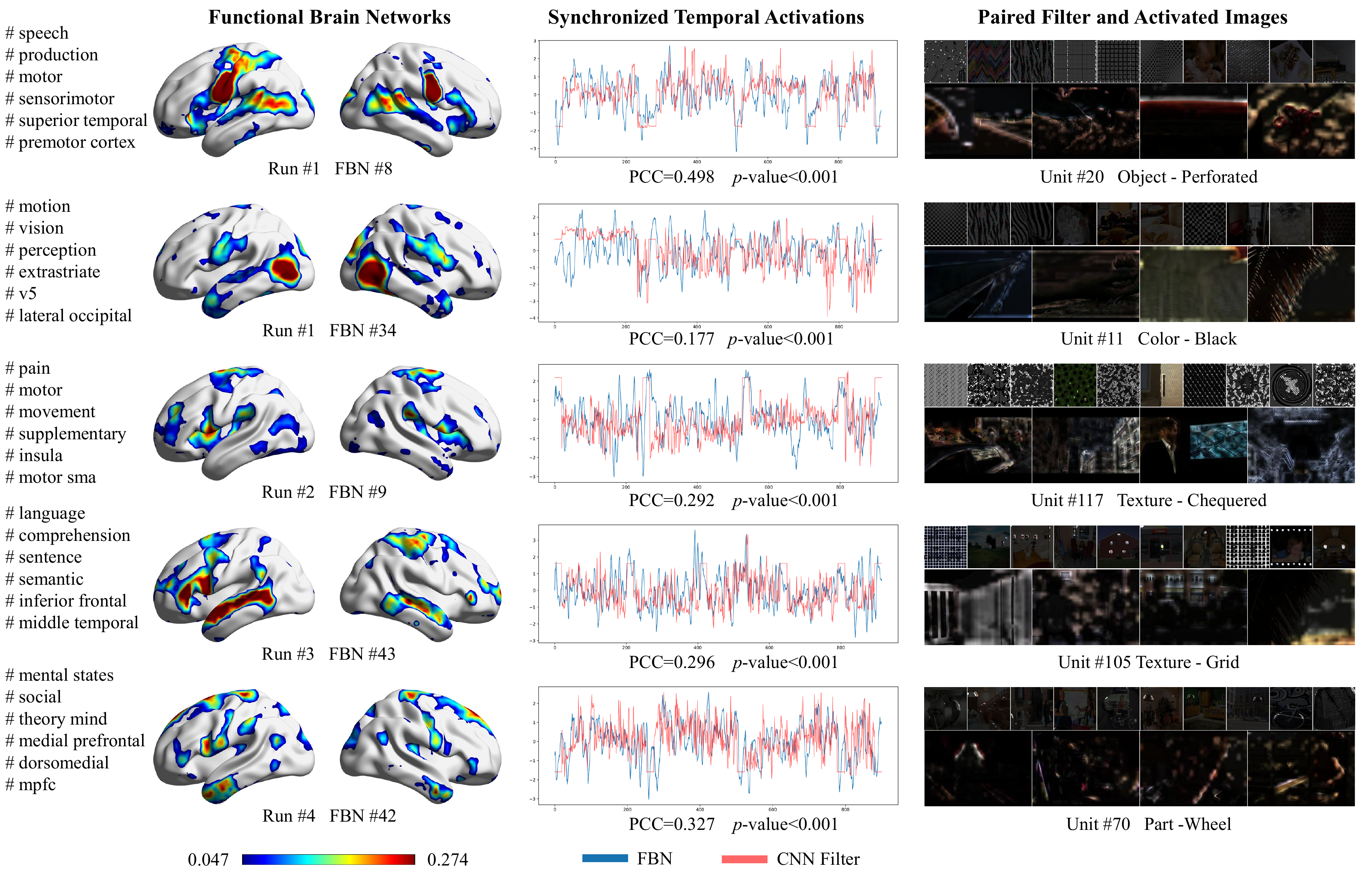}
  \caption{The visualization of FBN-Filter pairs obtained from ResNet-18 Block \#2. The left panel is the FBNs to be paired and semantic description from fMRI meta-analysis. The middle panel shows the synchronized activations from FBN and paired CNN filter. The right panel shows the most activated frames and the corresponding semantic description and filter's representative images~\cite{bau2017network}.}
  \label{figure2}
\end{figure}

\begin{figure}[h!]
  \centering
  \includegraphics[width=0.9\linewidth]{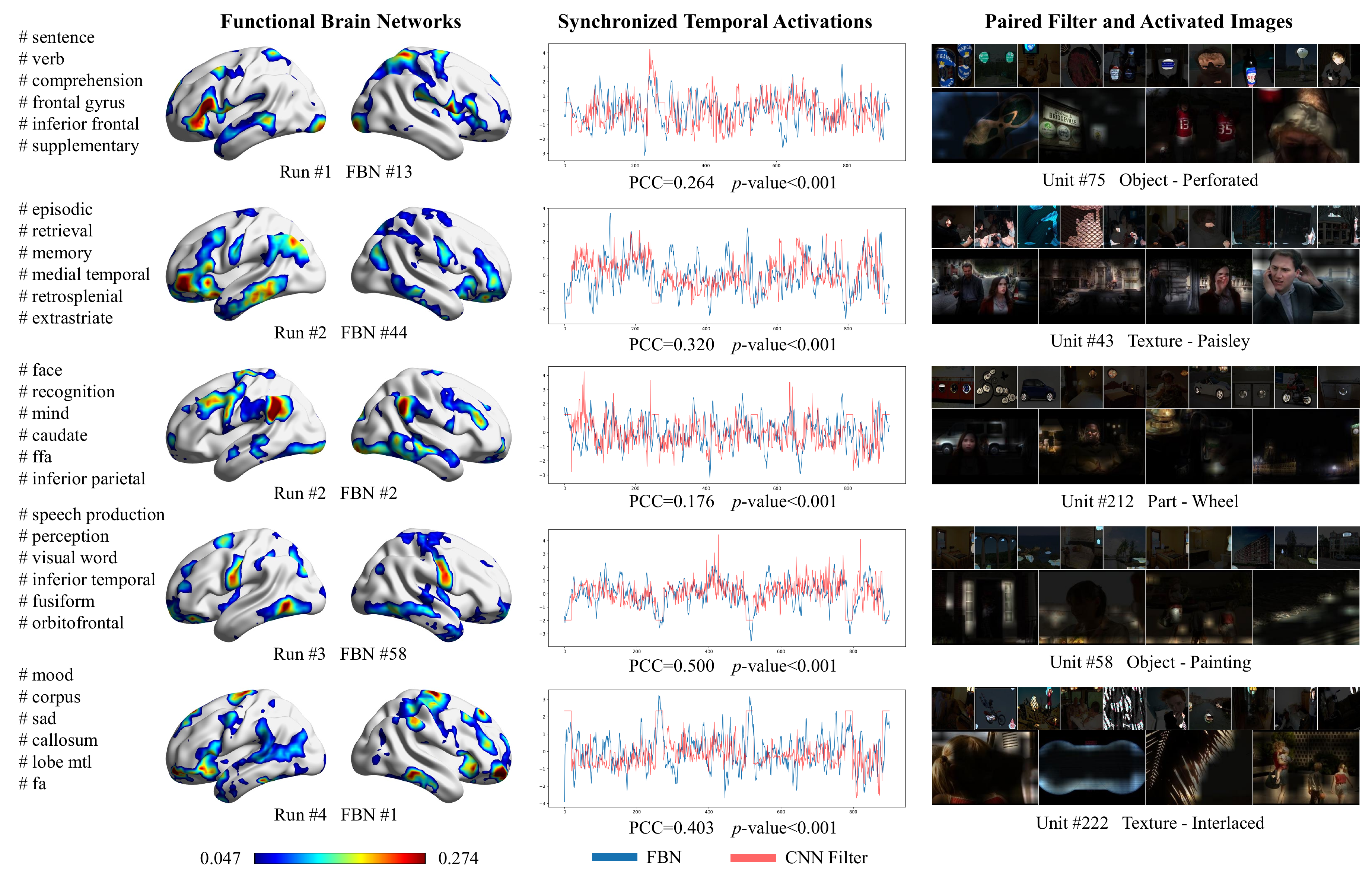}
  \caption{The visualization of FBN-Filter pairs obtained from ResNet-18 Block \#3. The left panel is the FBNs to be paired and semantic description from fMRI meta-analysis. The middle panel shows the synchronized activations from FBN and paired CNN filter. The right panel shows the most activated frames and the corresponding semantic description and filter's representative images~\cite{bau2017network}.}
  \label{figure3}
\end{figure}

\section{Visualizations of Cross-Annotation with Different Models}

In this section, we provide the examples for the cross-annotation based on the Eq. (2) to pair each FBN with a filter at the last convolutional layer of ResNet-50~\cite{he2016deep} and DenseNet-161~\cite{huang2017densely} model, and visualize several sample pairs in the Fig.~\ref{figure4} and Fig.~\ref{figure5}.

\begin{figure}[h!]
  \centering
  \includegraphics[width=0.9\linewidth]{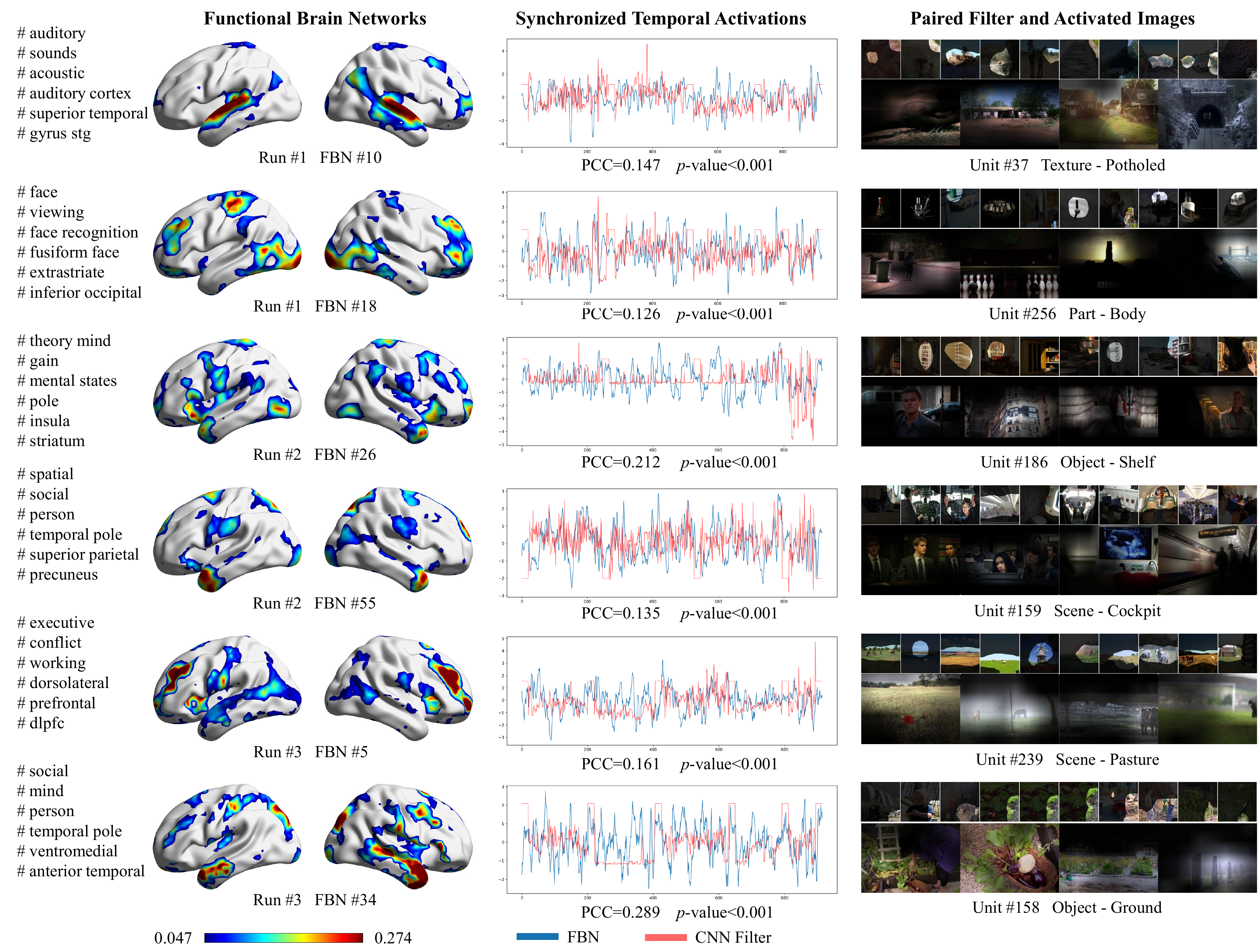}
  \caption{The visualization of FBN-Filter pairs obtained from the last convolutional layer of ResNet-50~\cite{he2016deep}. The left panel is the FBNs to be paired and semantic description from fMRI meta-analysis. The middle panel shows the synchronized activations from FBN and paired CNN filter. The right panel shows the most activated frames and the corresponding semantic description and filter's representative images~\cite{bau2017network}.}
  \label{figure4}
\end{figure}

\begin{figure}[h!]
  \centering
  \includegraphics[width=0.9\linewidth]{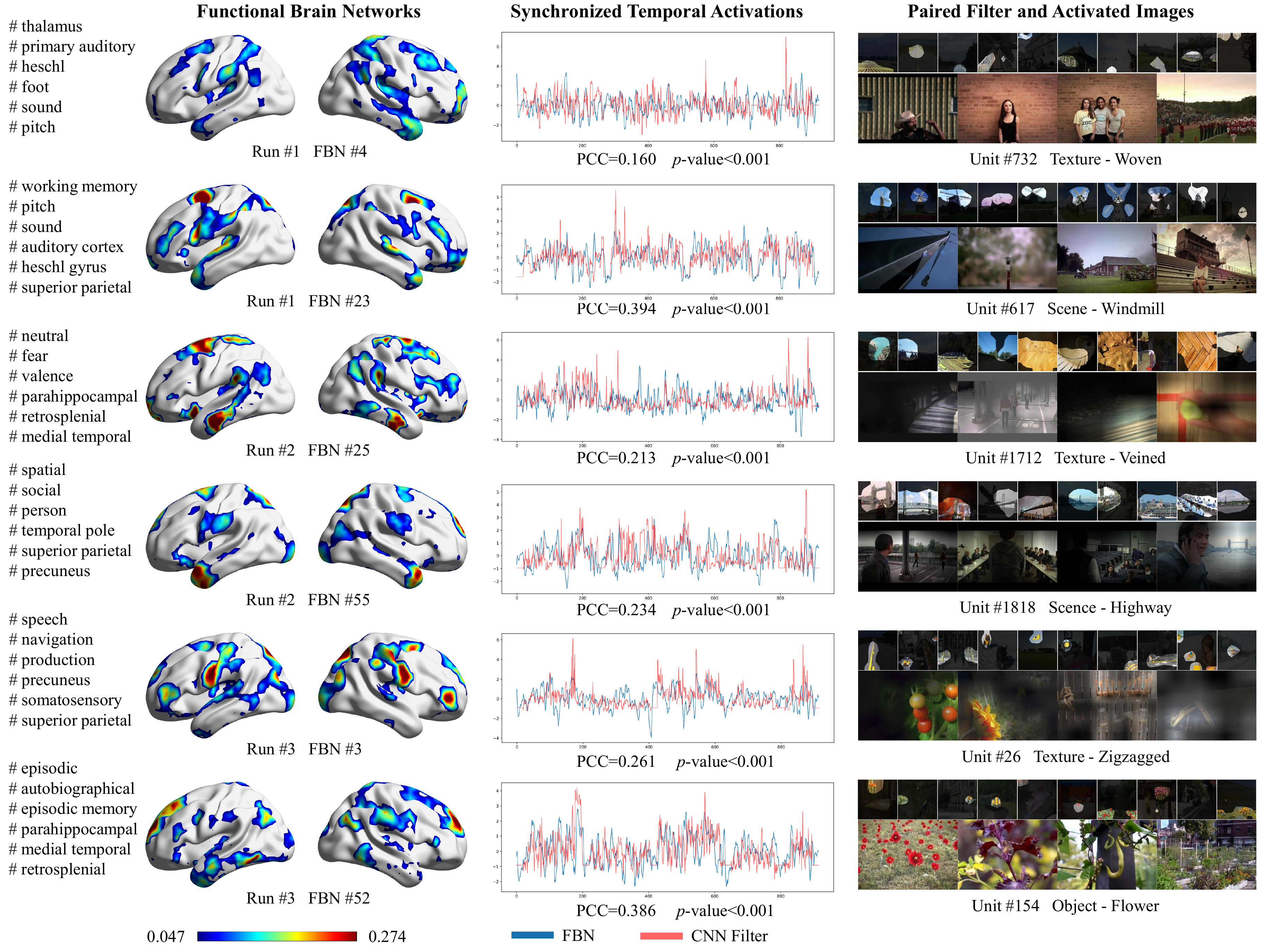}
  \caption{The visualization of FBN-Filter pairs obtained from the last convolutional layer of DenseNet-161~\cite{huang2017densely}. The left panel is the FBNs to be paired and semantic description from fMRI meta-analysis. The middle panel shows the synchronized activations from FBN and paired CNN filter. The right panel shows the most activated frames and the corresponding semantic description and filter's representative images ~\cite{bau2017network}.}
  \label{figure5}
\end{figure}

\section{Cross-Annotation based on Eq. (1)}
The experiments of our work are mainly based on the Eq. (2) to pair each FBN with a convolutional filter. In this section, we firstly report the averaged PCC in Table~\ref{table4} based on Eq. (1) to pair each covolutional filter with a FBN.

\begin{table}[ht]
\centering
\setlength{\tabcolsep}{1.25mm}
\caption{The averaged PCC across filters on HCP 7T movie-watching fMRI dataset for the pairs of filters in the last convolutional layers and FBNs on CNN models pre-trained on ImageNet and Places365 dataset. The PCC value in this table with a * marker indicates that several pairs (less than 10) are not statistically significant$(p$-value$ \leq 0.05)$, otherwise it is significant.}
\label{table4}
\begin{tabular}{lcccccccc}
\toprule

\multicolumn{1}{c}{\multirow{2}{*}{Methods}} & \multicolumn{4}{c}{ImageNet~\cite{deng2009imagenet}} & \multicolumn{4}{c}{Places365~\cite{zhou2017places}}\\
&  Run \#1 & Run \#2 & Run \#3 & Run \#4
&  Run \#1 & Run \#2 & Run \#3 & Run \#4\\ 
\midrule
AlexNet~\cite{krizhevsky2012imagenet} &0.2497 &0.2216 &0.2643 &0.2618 
&0.2581 &0.2366 &0.2706 &0.2602\\
VGG-16~\cite{simonyan2014very} &0.2191 &0.1944 &0.2267 &0.2253 & - & -&-&-\\
ResNet-18~\cite{he2016deep} &0.2467 &0.2230 &0.2533 &0.2415
&0.2210 &0.2177 &0.2392 &0.2358 \\
ResNet-50~\cite{he2016deep} &0.2362 &0.2233 &0.2462 &0.2378
&0.2428* &0.2281 &0.2511 &0.2424\\
DenseNet-161~\cite{huang2017densely} &0.2476 &0.2326 &0.2667 &0.2646 
&0.2438* &0.2294* &0.2517* &0.2472*\\
Inception V3~\cite{szegedy2016rethinking} &0.1910* &0.1901* &0.2026* &0.2003 & - & -&-&- \\
ShuffleNet V2~\cite{ma2018shufflenet} &0.2394* &0.2112* &0.2467* &0.2338* & - & -&-&- \\
MobileNet V2~\cite{sandler2018mobilenetv2} &0.2500 &0.2375 &0.2633 &0.2423 & - & -&-&- \\
ResNeXt-50~\cite{xie2017aggregated} &0.2408 &0.2162* &0.2449 &0.2297 & - & -&-&- \\
MNASNet~\cite{tan2019mnasnet} &0.2281 &0.2126 &0.2394 &0.2290 & - & -&-&- \\

\bottomrule
\end{tabular}
\end{table}

\begin{figure}[h!]
  \centering
  \includegraphics[width=0.9\linewidth]{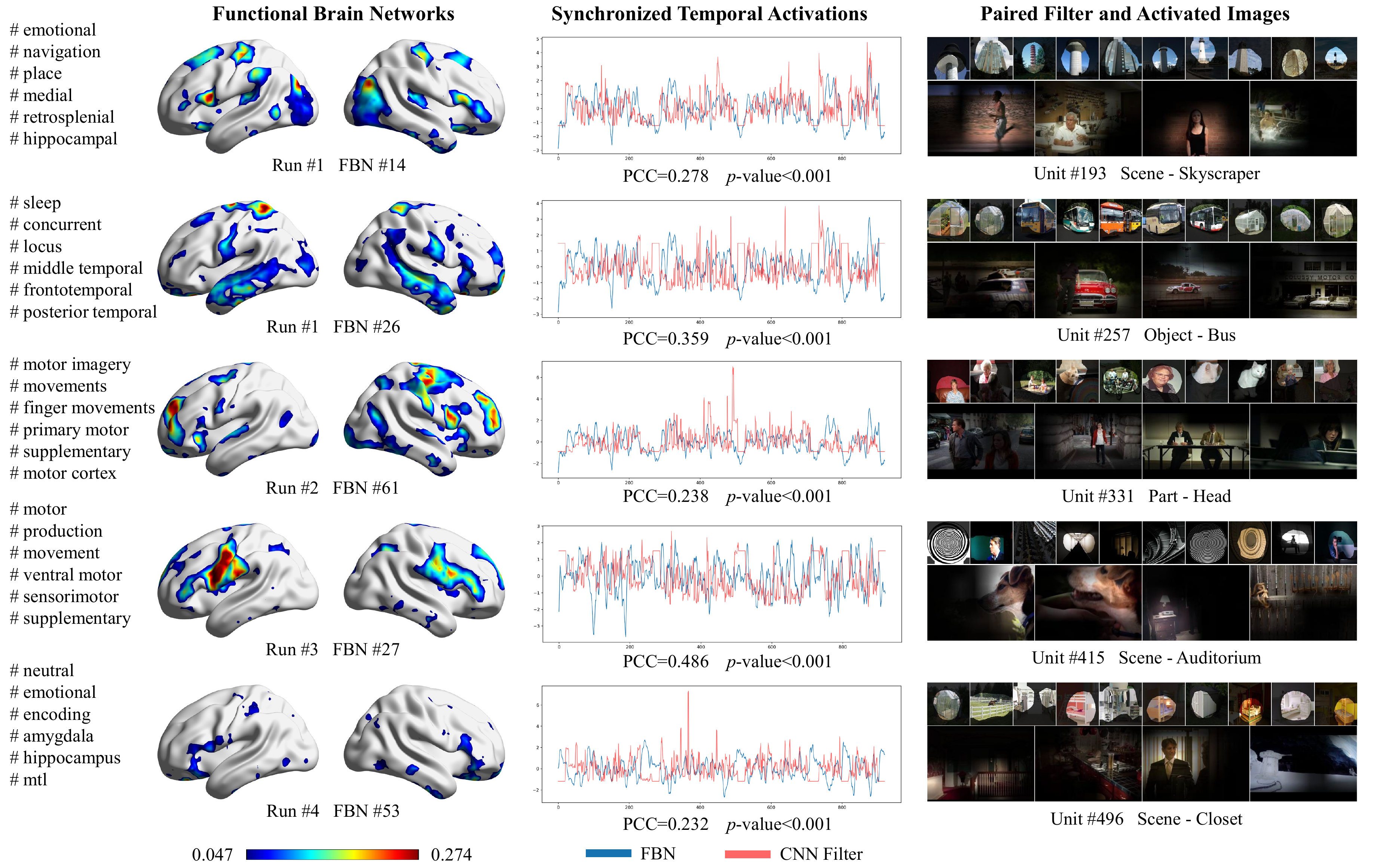}
  \caption{The visualization of Filter-FBN pairs obtained from the last convolutional layer of ResNet-18 model. The left panel is the paired FBNs and semantic description from fMRI meta-analysis. The middle panel shows the synchronized activations from CNN filter and paired FBN. The right panel shows the most activated frames and the corresponding semantic description and filter's representative images ~\cite{bau2017network}.}
  \label{figure6}
\end{figure}

We also randomly select several convolutional filters from the last layer of ResNet-18 model and visualize the paired FBNs in the Fig.~\ref{figure6}.

\section{Comparison of FBNs in Ablation Studies}

\begin{figure}[h!]
  \centering
  \includegraphics[width=0.9\linewidth]{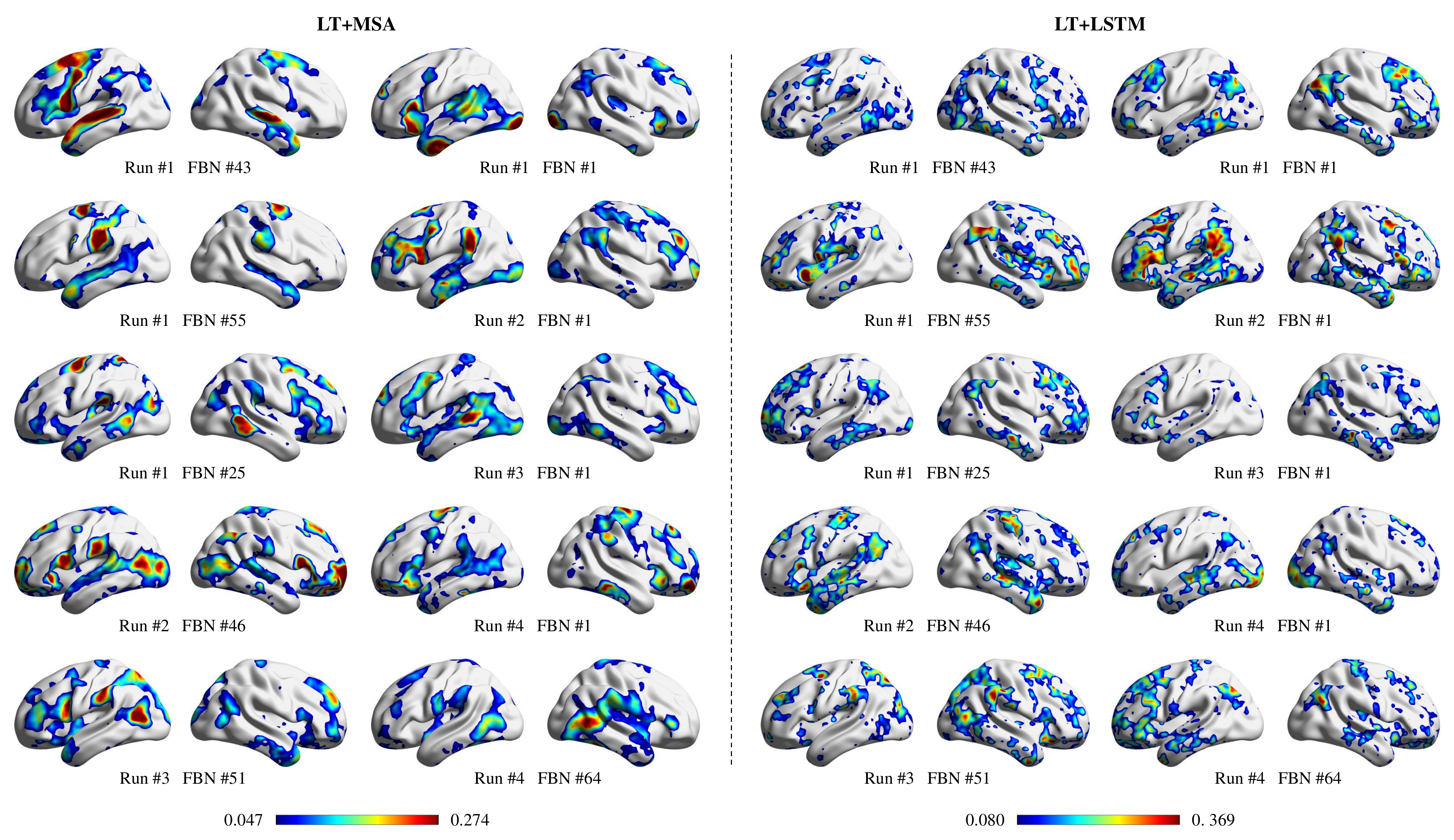}
  \caption{The visualization of FBNs generated from LT+MSA and LT+LSTM. Only the voxels between the $40^{th}$ percentile and $99^{th}$ percentile on the averaged histogram are shown for better visualization.}
  \label{figure7}
\end{figure}

In this section, we select 10 FBNs from LT+MSA and LT+LSTM model and visualize them in Fig.~\ref{figure7} for qualitative comparison. It is observed that FBNs obtained from LT+LSTM demonstrate a scattered distribution while those from LT+MSA are more concentrated on several meaningful regions. It is noted that such observation is well reproduced with other FBNs. So in our experiments and analyses, we used the FBNs obtained from LT+MSA model.

\end{document}